%% file: main-arxiv.tex
\documentclass{article}

\usepackage{iclr2027_conference,times}
\iclrfinalcopy
\usepackage{graphicx}
\usepackage{xcolor}
\usepackage{tikz}
\usetikzlibrary{arrows.meta}
\usepackage{float}
\usepackage{subcaption}
\usepackage{wrapfig}
\usepackage{amsthm}
\usepackage{amsmath}
\usepackage{amsfonts}
\usepackage[hypertexnames=false,hidelinks]{hyperref}
\usepackage{url}
\usepackage{thm-restate}
\usepackage{booktabs}
\usepackage{array,longtable}

\newcommand{\beq}{\begin{eqnarray}}
\newcommand{\eeq}{\end{eqnarray}}
\newcommand{\ba}{\begin{eqnarray*}}
\newcommand{\ea}{\end{eqnarray*}}

\makeatletter
\@ifundefined{newcounteralias}{}{\renewcommand\thmt@autorefsetup{\@xa\def\csname\thmt@envname autorefname\@xa\endcsname\@xa{\thmt@thmname}}}
\makeatother

\usepackage{natbib}
\newtheorem{theorem}{Theorem}[section]
\newtheorem{corollary}[theorem]{Corollary}
\newtheorem{lemma}[theorem]{Lemma}
\newtheorem{proposition}[theorem]{Proposition}
\theoremstyle{definition}
\newtheorem{definition}[theorem]{Definition}
\newtheorem{example}[theorem]{Example}
\theoremstyle{remark}
\newtheorem*{remark}{Remark}

\newcommand{\dout}{{d_\text{out}}}

\newcommand{\R}{\mathbb R}

\newcommand{\dd}{\mathrm d}

\newcommand{\sfA}{\mathsf{A}}

\newcommand{\sfE}{\mathsf{E}}

\newcommand{\sfG}{\mathsf{G}}

\newcommand{\sfM}{\mathsf{M}}

\newcommand{\sfV}{\mathsf{V}}

\newcommand{\sfX}{\mathsf{X}}
\newcommand{\sfY}{\mathsf{Y}}

\newcommand{\calD}{\mathcal{D}}

\newcommand{\calK}{\mathcal{K}}

\newcommand{\calN}{\mathcal{N}}

\newcommand{\calP}{\mathcal{P}}

\newcommand{\calR}{\mathcal{R}}
\newcommand{\calS}{\mathcal{S}}
\newcommand{\calT}{\mathcal{T}}

\newcommand{\identity}{\mathrm{id}}
\DeclareMathOperator{\supp}{supp}
\DeclareMathOperator*{\argmin}{arg\,min}

\title{All roads lead to Markov kernel networks: A representation theorem for discretization-stable learning}
\author{\begin{tabular*}{\dimexpr\textwidth-2\tabcolsep\relax}[t]{@{}l@{\extracolsep{\fill}}ll@{}}
        \textbf{S David Mis} & \textbf{Trevor Teolis} & \textbf{Ivan Dokmani\'{c}} \\
        \textnormal{Rice University} & \textnormal{Rice University} & \textnormal{University of Basel} \\[1.5ex]
        \textbf{Matti Lassas} & \textbf{Maarten V. de Hoop} & \\
        \textnormal{University of Helsinki} & \textnormal{Rice University} &
    \end{tabular*}
}

\date{\today}

\begin{document}

\maketitle
\lhead{Preprint}
\begin{abstract}
    Operator learning seeks to approximate function-to-function maps in a way that remains stable when the input discretization changes, a property sometimes called ``discretization invariance''.
    We consider a quantitative refinement of this idea, called \emph{uniform discretization stability}, which builds on a common representation of input functions and their discretizations as probability measures and provides a H\"older bound relating output changes to the Wasserstein distance between these measures.
    Our main result is a representation theorem showing all uniformly discretization stable operators can be written exactly as a \emph{stable Markov kernel network}---an abstract architecture that is a joint generalization of transformers, Fourier neural operators (FNO), Flowers, and other successful learned operators. Our representation theorem characterizes a design space for discretization-stable operator learning; designing new architectures becomes a question of which parts of a Markov kernel network to parameterize, and how. We further show that the read kernels of these networks can be realized as solutions of regularized variational problems, with particular choices of costs and regularizers recovering the kernels of transformers, FNO, Flowers, their hybrids, and beyond. This variational formulation provides a systematic approach to architecture design, together with sufficient conditions for uniform discretization stability.
\end{abstract}

\section{Introduction}

Operator learning requires neural networks that approximate maps between functions. The input functions of the target operator are continuum objects that cannot be represented exactly on a computer. Furthermore, in scientific applications using real-world data, we generally do not have complete knowledge of the input function $f$, but rather only a finite collection of (sometimes noisy) samples from the graph of a function $(x, f(x)).$ To accurately model the continuum target operator, a neural network should not be sensitive to small changes in the discretization of the input. The output of a  ``discretization invariant'' network converges as the discretization of the input function is refined.
Quantitative guarantees about the trained network, however, require stability bounds, not merely convergence under discretization refinement. This leads to the following three questions:
\begin{enumerate}
    \item What discretization stability properties are relevant for operator learning?
    \item What network architectures satisfy these stability properties?
    \item How can these architectures be characterized and systematically explored?
\end{enumerate}

Our approach is to lift operators between functions to operators between measures. This provides a unified viewpoint of continuum inputs (functions) and their noisy discretizations. The necessary stability properties can then be stated as a H\"older stability with respect to a Wasserstein metric. \emph{Uniform discretization stability} controls the behavior of learned operator networks under different discretizations of the input functions. For example, it provides quantitative guarantees for networks that are trained at one resolution or on a uniform grid, but evaluated on non-uniform point clouds sampled at a different resolution. It also provides a unified framework for the analysis of errors arising from both discretization and noisy measurements.

We then turn to the second question, and introduce \emph{Markov kernel networks}, a shared generalization of transformers \citep{vaswani2017attention}, FNO \citep{li2021fourier}, Flowers \citep{muser2026flowers}, and other operator networks. These networks alternate nonlinear integral operators built using Markov kernels to mix information spatially with channel-wise mixing operators to mix information pointwise in the embedding space. When the individual components of these networks satisfy relevant stability properties, the entire network becomes uniformly discretization stable. Conversely, our main theorem shows that every uniformly discretization stable operator can be written exactly (not approximately) as a stable Markov kernel network.

With the design space of discretization-stable architectures fixed by the representation theorem, we propose a variational approach to explore the design space. We investigate Markov kernels that minimize regularized objectives; the induced read behavior is determined by the choice of regularizer and cost function. We provide sufficient conditions on the regularizers and costs so that the resulting networks are guaranteed to be uniformly discretization stable, and identify the regularizers corresponding to the popular network architectures discussed earlier. We aim to provide a principled framework for adaptive \citep{cortes2017adanet} and hybrid \citep{balaji2026hybrid} architectures, and a useful belief space for agentic \citep{toscano2026athenaagenticteamhierarchical} approaches to operator learning.

\subsection{Prior work}

Discretization invariance was initially explored in the context of ill-posed inverse problems. The term was first defined in \citet{lassas2004totalvariation}, with important antecedents in \citet{lasanen2002discretizations,dambrogi1999discretization,lehtinen1989linear}; see \cite{lassas2009discretization} and \citet{stuart2010inverse} for additional historical remarks.
It was popularized in operator learning by (earlier preprints of) the seminal work \citet{kovachki2023neuraloperator}, which includes analysis of operators that converge under discretization refinement. Various notions of quantitative discretization invariance and stability bounds have been developed in \citet{wang2023discretization,bartolucci2023representation,lanthaler2024discretization,debie2019stochastic,ruiz2020graphon,gao2025discretization,subedi2025controlling}, and \citet{bendahi2026discretization}.
Stability of transformers was investigated in \citet{castin2024smooth}.
The notion of uniform discretization stability we propose in Section \ref{sec:uniform-discretization-stability} has roots in Transportation-$L^p$ theory \citep{garciatrillos2016continuum,thorpe2017transportation}; we are unaware of prior work using these H\"older bounds in the context of operator learning, but \cite{furuya2026functiongraph} discusses a similar Lipschitz bound for transformers. This paper builds on both the stability bound and ``graph measure'' viewpoint of that work.

The mindset for our representation theorem and variational characterizations is inspired by \citet{unser2019representer}, which proved that optimal activation functions (in a total-variation sense) for multi-layer perceptrons take the form of linear splines with adaptive knot locations. Our presentation of Markov kernel networks directly mirrors measure-theoretic transformers \citep{furuya2024transformersInContext,furuya2025support,furuya2026functiongraph}. \citet{niculae2017regularized} characterized softmax attention as a minimizer of a KL-regularized objective, which was further explored in \citet{martins2022sparsecontinuous,terjek2021moreau}.

\subsection{Paper outline and contributions}

\begin{itemize}
    \item In Section \ref{sec:uniform-discretization-stability}, we propose \emph{uniform discretization-stability}, a H\"older stability property, as a quantitative notion of discretization invariance.
    \item In Section \ref{sec:MKN}, we define \emph{Markov kernel networks}, an abstract architecture that generalizes transformers, neural operators, Flowers, and others.
    Our main results are the representation theorems \ref{thm:representation-1} and \ref{thm:representation-2} establishing an equivalence between uniformly discretization-stable operators and stable Markov kernel networks.
    This suggests Markov kernel networks provide a natural design space for uniformly discretization-stable operator learning.
    \item Section \ref{sec:variational-principles} explores the design space of Markov kernel networks using a variational characterization of discretization stable networks. We investigate Markov kernels that minimize a particular regularized objective function, and show how certain regularizers determine different read behavior, including those corresponding to existing architectures and hybrids. We establish stability bounds for variationally defined Markov kernels.

\end{itemize}
Section \ref{sec:discussion} discusses some limitations of our results. The appendices contain background and notation used throughout the document (Appendix \ref{sec:notation-background}), proofs (Appendix \ref{sec:proofs}), a universal approximation result for Markov kernel networks (Appendix \ref{sec:universal-approximation}), constructions for popular architectures as Markov kernel networks (Appendix \ref{sec:concrete-subclasses}),  and an additional discussion of hybrid architectures defined by variational Markov reads (Appendix \ref{app:s4-proofs}).

\section{Uniform discretization stability}
\label{sec:uniform-discretization-stability}

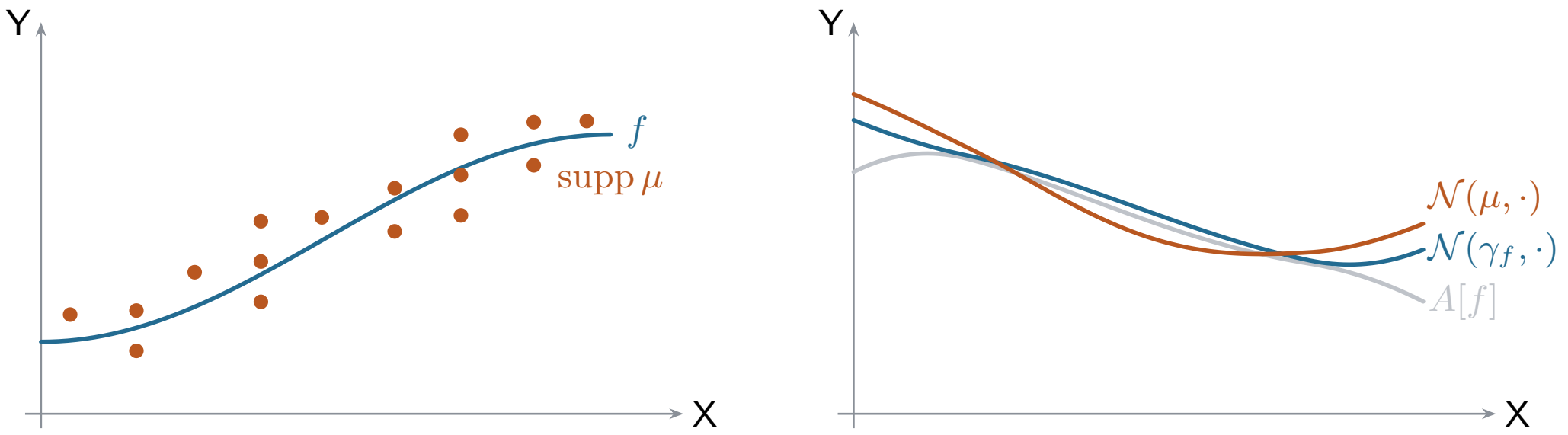
\begin{figure}[t]
    \centering
    \begin{subfigure}[t]{0.495\linewidth}
        \centering
        \input{figures/operator-learning-input.tex}
        \caption{Inputs}
        \label{fig:operator-learning-input}
    \end{subfigure}\hfill
    \begin{subfigure}[t]{0.495\linewidth}
        \centering
        \input{figures/operator-learning-output.tex}
        \caption{Outputs}
        \label{fig:operator-learning-output}
    \end{subfigure}
    \caption{Operator learning setup. The target operator is a continuum object $f \mapsto A[f]$. A neural network $\calN$ takes as input a measure $\mu$ representing a noisy sample of points from the graph of $f$ (denoted $\gamma_f$), and outputs a function that can be evaluated at any point in the domain $\sfX$. The network $\calN$ is \emph{uniformly discretization stable} (Definition \ref{def:DS}) if it satisfies a H\"older stability property that provides quantitative guarantees that $\calN(\mu, \cdot)$ converges to $\calN(\gamma_f, \cdot)$, the output had the network been provided with the ideal (continuum) input $\gamma_f$.
    }
    \label{fig:operator-learning-setup}
\end{figure}

The target operator maps functions to functions, but in practice the network receives samples of the input function and their locations (Figure~\ref{fig:operator-learning-setup}). We represent both these samples and the continuum input as probability measures on a space of functions graphs \citep{furuya2026functiongraph}. The network then maps an input measure to an output function, allowing us to quantify how changes in sampling affect its predictions.

Let $(\sfX, \rho_\sfX)$ be a compact metric space with at least two elements,
and $\sfY \subset \mathbb{R}^\dout$ be non-empty, compact, and convex. Define the \emph{graph metric space} $\sfG := \sfX \times \sfY$ equipped with the metric
\begin{equation}
    \rho_\sfG: \sfG \times \sfG \to \mathbb{R}_{\ge 0}, \qquad \rho_\sfG((x_1, y_1), (x_2, y_2)) := \sqrt{ \rho_\sfX(x_1, x_2)^2 + |y_1 - y_2|^2}.
\end{equation}
Choose a Radon probability measure $\lambda \in \calP(\sfX)$ with $\supp \lambda = \sfX$. Let $L^2(\sfX, \sfY)$ be the space of Radon-measurable functions $f:\sfX\to \sfY$ that are square integrable with respect to $\lambda$.
Observe that since $ \sfY$ is compact,  $L^2(\sfX, \sfY)\subset  L^\infty(\sfX, \sfY)$.
In this paper, we consider operators $A: L^2(\sfX, \sfY) \to L^2(\sfX, \sfY)$. All maps in the paper are Borel measurable.

We identify a function $f \in L^2(\sfX, \sfY)$  with its \emph{graph measure} \citep{furuya2026functiongraph},
\begin{equation}
    \label{graph-measure}
    \gamma_f \in \calP(\sfG), \qquad \gamma_f := (\identity, f)_\# \lambda,\quad (\identity, f):\sfX\to \sfX\times \sfY.
\end{equation}
(See Appendix \ref{sec:notation-background} for the definition of the pushforward operator ``$\phantom{}_\#$''.) The map $f \mapsto \gamma_f$ is injective.
For helpful intuition, when $f$ is continuous (so it is defined pointwise and has a closed graph) we have $\supp \gamma_f = {\{ \big(x, f(x)\big) : x \in \sfX \}}$.

Computer implementations of neural networks for function-to-function maps generally do not accept literal functions $f \in L^2(\sfX, \sfY)$, but instead $f$ is approximated by a finite collection of points on $\gamma_f$, or only near $\gamma_f$ if measurements include noise.
We model these sampled points as a mixture of Dirac measures.
Numerically, $f \in L^2(\sfX, \sfY)$ is approximated by sampling a finite collection of domain points $\{x_i\}_{i=1}^N$ and forming an empirical approximation of $\gamma_f$,
\begin{equation}
    \mu \in \calP(\sfG),
    \qquad
    \mu = \sum_{i=1}^N w_i \delta_{(x_i, y_i)}, \quad w_i \ge 0, \quad \sum_{i=1}^N w_i = 1.
\end{equation}
When the samples $(x_i, y_i)$ lie on the graph of a function $f$, we have $y_i = f(x_i)$.
The mixture weights $w_i$ allow the samples to approximate a graph measure even when the $x_i$ locations are not distributed uniformly with respect to $\lambda$. This formulation captures both discretizations on uniform grids and irregularly sampled point clouds. Provided with samples $\mu$ to represent the input function $f$, an operator network $\calN: \calP({\sfG}) \times \sfX \to \sfY$ can then be evaluated at an arbitrary $x \in \sfX$.
Informally, a trained neural network $\calN$ is a good approximation of $A$ when
\begin{equation}
    \label{eq:approximation_criterion}
    \mu \approx \gamma_f \;\Longrightarrow\; \calN(\mu, \cdot) \approx A[f] \qquad \text{for all } f \in L^2(\sfX, \sfY).
\end{equation}

To make the above precise, we require a metric on measures to compare the graph measure $\gamma_f$ to the noisy samples $\mu$.
For this, we use the Wasserstein-1 optimal transport metric (Appendix \ref{def:wasserstein-1}.)

We now formulate \emph{uniform discretization stability} as a H\"older-stability property for neural networks.

\begin{definition}[Uniform discretization stability]
\label{def:DS}
A network $\calN: \calP({\sfG}) \times \sfX \to \sfY$ is \emph{uniformly discretization stable} on a set $\calS \subseteq \calP(\sfG)$
if there exist constants $C > 0$, $\beta \in (0, 1]$ such that
\begin{equation}
    \label{eq:DS}
    \tag{DS}
    \sup_{x\in\sfX} | \calN(\mu,x) - \calN(\nu,x) |
    \le C\, W_1(\mu, \nu)^\beta \qquad \text{for all } \mu, \nu \in \mathcal{S}.
\end{equation}

\end{definition}
Returning now to the approximation criterion \eqref{eq:approximation_criterion}, using the triangle inequality and the fact $\lambda(\sfX) = 1$,
\begin{equation}
    \label{eq:approximation-error-decomposition}
    \| \mathcal{N}(\mu, \cdot) - A[f] \|_{L^2(\sfX, \mathbb{R}^\dout)}
    \;\le\;
    \underbrace{\| \mathcal{N}(\gamma_f, \cdot) - A[f] \|_{L^2(\sfX, \mathbb{R}^\dout)}}_{\text{ideal approximation error}}
    \;+\;
     \underbrace{\| \mathcal{N}(\mu, \cdot) - \mathcal{N}(\gamma_f, \cdot) \|_\infty}_{\text{discretization error}}.
\end{equation}
The first term on the right is the approximation error of an ideal network given all information about the input function $f$; this quantity is controlled by training the network.
The second term is the error introduced by the discretization $\mu$.
It is largely determined by the neural network architecture itself, and is not directly controlled by training.
Discretization stable architectures satisfying \eqref{eq:DS} control the discretization error, so they have the possibility to be trained to satisfy the desired property \eqref{eq:approximation_criterion}.
In the next section, we will show that all such networks have a particular representation as stable Markov kernel networks.
This allows us to view, for example, softmax attention, and SelfWarps \citep{muser2026flowers} as two particular choices in the same space of discretization stable architectures.

\begin{remark}
    Uniform discretization stability is a stronger condition than convergence under resolution refinement.
    In particular, \eqref{eq:DS} allows a unified analysis of both resolution refinement and stability to noise in the measurements of $f$, both of which are important in applications.
    Furthermore, for sufficiently regular classes of target operators, it allows us to obtain quantitative guarantees for the performance of a network trained at one resolution but evaluated at another, which is important for scientific applications where large quantities of high-resolution training data may be prohibitively expensive. Indeed, if $\mathcal{N}$ is uniformly discretization stable, then
    \begin{equation}
        \| \mathcal{N}(\mu_{\text{test}}, \cdot) - A[f] \|_{L^2(\sfX, \mathbb{R}^\dout)} \le \| \mathcal{N}(\mu_{\text{train}}, \cdot) - A[f] \|_{L^2(\sfX, \mathbb{R}^\dout)}
        +
        C\, W_1(\mu_{\text{test}},\mu_\text{train})^\beta,
    \end{equation}
    where $\mu_{\text{train}}$ and $\mu_{\text{test}}$ are discretizations of $f$ at the training and test resolutions, respectively.
    Simple convergence under resolution refinement provides no such guarantees.
\end{remark}

\begin{remark}
    Definition \ref{def:DS} does not impose additional regularity requirements of the target operator $A$ or the input functions $f$. See the discussion of universal approximation in Appendix \ref{sec:universal-approximation}.
\end{remark}

\subsection{Non-collapsing measures $\calS$}

\begin{wrapfigure}[9]{r}{0.32\textwidth}
    \centering
    \vspace*{-0.8cm}
    \input{figures/non-admissible-measure.tex}
    \captionsetup{font=footnotesize,skip=4pt}
    \caption{A measure that is not in any admissible set $\calS$. Since $(\pi_\sfX)_\# \mu = \delta_a$, every set containing $\mu$, or a sequence converging to $\mu$, fails the admissibility criterion \eqref{eq:admissibility-criterion}.}
    \label{fig:non-admissible-measure}
\end{wrapfigure}
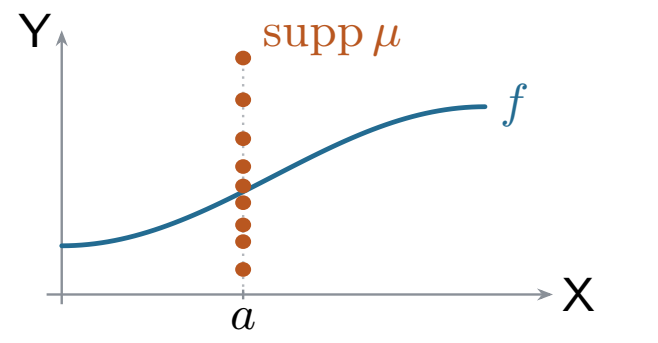

The set $\cal{S} \subset \calP(\sfG)$ in Definition \ref{def:DS} need not contain all probability measures to provide a useful notion of discretization stability. Our main result, Theorem \ref{thm:representation-2}, holds for admissible classes that are relevant for operator learning, but not all possible inputs in $\calP(\sfG)$ (see Example \ref{example:counterexample}).

A set of measurements at only a single value of $x$ does not provide much information about an input function $f$, and we are not concerned with the behavior of networks on such pathological inputs. To avoid this pathology, we define a condition on admissible sets of measures to ensure there is enough variation in the $\sfX$-marginal.

\WFclear
\begin{definition}[Admissible sets of non-collapsing measures]
    \label{def:non-collapsing-measures}
    A non-empty set $\calS \subset \calP(\sfG)$ is \emph{admissible} if there exists $\epsilon>0$ such that
    \begin{equation}
        \label{eq:admissibility-criterion}
        \inf_{\mu \in \calS} \inf_{a \in \sfX} W_1 \big( (\pi_\sfX)_\# \mu, \delta_a \big) > \epsilon
    \end{equation}
    where $\pi_\sfX: (x, y) \mapsto x$ is projection onto the $\sfX$ component.
\end{definition}

Crucially, $\calS$ can contain both graph measures $\gamma_f$ and empirical measures $\mu$ built from finitely many samples, even in the presence of a large amount of noise, as long as the $x$ coordinates do not collapse.

\section{Markov kernel networks}
\label{sec:MKN}

We now define an abstract architecture for operator learning called \emph{Markov kernel networks}, and show they exactly represent all uniformly discretization stable operators on admissible input classes.

\subsection{Definitions}

Let $\sfM$ be a metric space. A nonlinear \emph{Markov kernel} is a map $K: \calP_1(\sfM) \times \sfM \to \calP_1(\sfM)$.
Markov kernels form the backbone of the ``in-context maps'' \citep{furuya2024transformersInContext} that define the hidden layers of a Markov kernel network. An in-context map uses a state encoded as a probability measure, called a ``context'', to determine how to update a single point.

\begin{definition}[Markov kernel in-context map]
    \label{def:markov-kernel-in-context-map}
    Let $\sfE$, $\sfE'$ be compact metric spaces.
    A \emph{Markov kernel in-context map} $\Gamma: \calP(\sfE) \times \sfE \to \sfE'$ with $H > 0$ heads is of the form
    \begin{align}
        \label{eq:read-head}
        \qquad r_h(\mu, z) &= \int_\sfE v_h(w) K_h[\mu,z]( \dd w),
        \qquad h = 1, \dots, H \\
        \label{eq:Gamma}
        \Gamma(\mu, z) &= m \big(z, r_1(\mu,z), \dots, r_H(\mu,z) \big).
    \end{align}
For $h = 1, \dots, H$, the maps $K_h: \calP(\sfE) \times \sfE \to \calP(\sfE)$ are Markov kernels, the bounded maps $v_h: \sfE \to \R^{d_h}$ are called \emph{value maps}, and $m: \sfE \times \prod_{h=1}^H \R^{d_h} \to \sfE'$ is called a \emph{channel mixer}. Each $r_h: \calP(\sfE) \times \sfE \to \R^{d_h}$ is called a \emph{read head}. Additionally, we require the output of the Markov kernels to be supported on the input context $\mu$:
        \begin{equation}
            \label{eq:markov_kernel_support}
            \supp K_h[\mu,z] \subset \supp \mu \qquad \text{for all } \mu \in \calP(\sfE),\; z \in \sfE,\; h = 1, \dots, H.
        \end{equation}
\end{definition}

\begin{remark}
The support condition \eqref{eq:markov_kernel_support} says that we may only access information provided in the current context, which is derived from the sampled points on the graph of the input function $f$. We have no access to information about $f$ outside the sampled points. The read heads mix information spatially across the embedding space $\sfE$; conversely, the channel-mixers are pointwise in $\sfE$.
\end{remark}

We use Markov kernel in-context maps to define neural network layers.
The input to the layer is a current context that was output from the previous layer, and was ultimately derived from the sampled points of the input function.
The layer updates the context by applying the in-context map to each point; that is, the context is pushed-forward by the in-context map.

\begin{definition}[Markov kernel layer]
    A Markov kernel layer $B: \calP(\sfE) \to \calP(\sfE')$ determined by the Markov kernel in-context map $\Gamma: \calP(\sfE) \times \sfE \to \sfE'$ is defined
    \begin{equation}
        B_\Gamma(\mu) = \Gamma(\mu, \cdot)_\# \mu.
    \end{equation}
\end{definition}

Appendix \ref{sec:attention-as-pushforward} contains an explicit construction of a softmax attention layer from a transformer using our notation to provide a link to a familiar setting.

We can now define the full Markov kernel network abstract architecture. A Markov kernel network $\calN: \calP(\sfG) \times \sfX \to \sfY$ encodes the input context into a lifted embedding space, updates the context using Markov kernel layers, then decodes the final context at a query point to obtain the output.

\begin{definition}[Markov kernel network]
    \label{def:MKN}
    Let $L > 0,$ and $\sfE_\ell$ be a compact metric space for $\ell=0, \dots, L$. A \emph{Markov kernel network} $\calN: \calP(\sfG) \times \sfX \to \sfY$ is of the form
    \begin{equation}
        \begin{aligned}
            \mu_0 &:= {q_\sfG}_\# \mu, \\
            \mu_L &:= (B_{L-1} \circ \dots \circ B_0)(\mu_0), \\
            \calN(\mu, x) &:= \Gamma_p \big(\mu_L, q_\sfX(x) \big),
        \end{aligned}
        \qquad\qquad
        \mu \in \calP(\sfG),\; x \in \sfX
    \end{equation}
    where
    \begin{itemize}
        \item $q_\sfG: \sfG \to \sfE_0$ and $q_\sfX: \sfX \to \sfE_L$ are pointwise encoders that lift inputs to hidden embedding spaces. We note that $q_\sfX=q_\sfX(x)$ is independent of the context $\mu.$
        \item $B_\ell: \calP(\sfE_\ell) \to \calP(\sfE_{\ell+1})$ are  Markov kernel layers for $\ell = 0, \dots, L-1$.
        \item $\Gamma_p: \calP(\sfE_L) \times \sfE_L \to \sfY$ is a decoder in the form of a Markov kernel in-context map.
    \end{itemize}
\end{definition}

Different concrete architectures are obtained by specifying the Markov kernels $K$, value maps $v$, and channel mixers $m$ (all of these for both the hidden context layers $B_\ell$ and the decoder $\Gamma_p$), as well as the encoders $q_\sfG$ and $q_\sfX$.
We give the example for a tranformer below; additional examples with full constructions (including, for example, assorted normalization operations) can be found in Appendix \ref{sec:concrete-subclasses}.

\begin{example}[Transformers]
    Consider the simple absolute positional encoding $z := (x, a) \in \mathbb{R}^{d}$, $d = d_x + d_a$, where $x \in \mathbb R^{d_x}$ and $a \in\mathbb R^{d_a}$ is the
    current hidden features.
    We preserve spatial coordinates $x$ in each token.
    Suppressing the layer index, each head $h=1,\ldots,H$ uses a softmax
    attention kernel
    \begin{equation}
        \begin{aligned}
            K_h[\mu,z](\mathrm dw)
            =\frac{\exp s_h(z,w)}
            {\displaystyle\int_{\mathsf E}
                \exp s_h(z,\widetilde w)\,\mu(\mathrm d\widetilde w)}
            \,\mu(\mathrm dw),
            \quad
            s_h(z,w)
            :=\frac{\langle Q_h z,K_{\mathrm{key},h} w\rangle}{\sqrt{k_h}}.
        \end{aligned}
    \end{equation}
    Here, $Q_h,K_{\mathrm{key},h} \in \mathbb R^{k_h\times d}$, $V_h \in \mathbb R^{d_a \times d}$ are query, key, and value matrices, respectively, and $k_h$ is the query/key dimension.
    Combining these $K_h$ with linear value maps $v_h(w):=V_h w$ yields each read head $r_h$ according to equation \eqref{eq:read-head}.
    The channel mixers include skip connections and feed-forward MLPs $F:\mathbb R^{d}\to\mathbb R^{d_a}$:
    \begin{equation}
        m\bigl((x,a),r_1,\ldots,r_H\bigr)
        =\Biggl(x, F \Bigl(x, a+\sum_{h=1}^{H}r_h \Bigr) \Biggr).
    \end{equation}
    This construction is a simplified function graph transformer layer \citep{furuya2026functiongraph}; other transformer variants for operator learning can be constructed in a similar manner.
\end{example}

Popular architectures can largely be classified by how information is routed through the network by their Markov kernels. Transformers are distinguished by their specific form of scaled dot-product attention. Nonlinear neural operators can be realized by multiple heads and a signed-measure decomposition of a nonlinear integral kernel $k(\mu, z, w)\,\mu(\dd w)$; Fourier neural operators \citep{li2021fourier} can be realized by the simple choice $K[\mu, z] = \mu$ with spectral computations occurring in $v$ and $m$. Flowers, on the other hand, uses Markov kernels that output a Dirac measure.
Furthermore, branch-and-trunk style networks, such as DeepONet \citep{chen1995operators,lu2021deeponet}, also fit into the Markov kernel network abstract architecture. The distinguishing features of these networks are the forms of their decoders $\Gamma_p$.

\subsection{A representation theorem for uniformly discretization-stable operators}

We now define \emph{stable Markov kernel networks} as the subset of Definition \ref{def:MKN} whose individual components all satisfy certain stability properties, and show this guarantees the entire network is uniformly discretization stable. We then state our main result that every operator that is uniformly discretization stable on a class of input measures satisfying Definition \ref{def:non-collapsing-measures} is exactly representable as a stable Markov kernel network.

\begin{definition}[Stable Markov kernel network]
    \label{def:stable-markov-kernel-network}
    A Markov kernel network $\calN: \calP(\sfG) \times \sfX \to \sfY$ is \emph{stable} if the following hold:
    \begin{enumerate}
        \item The encoders $q_\sfG$, $q_\sfX$, value maps $v_h$, and channel mixers $m$ are Lipschitz.
        \item In every hidden layer $B_\ell$, every Markov kernel $K_h: \calP(\sfE_\ell) \times \sfE_\ell \to \calP(\sfE_\ell)$ is H\"older stable in both arguments: there exists $C > 0$ and $\beta \in (0,1]$ such that
              \begin{equation}
                  \label{eq:joint-stability-1}
                  W_1\big(K_{h}[\mu,w], K_{h}[\nu,z] \big) \le C \big( W_1(\mu, \nu) + \rho_{\sfE_\ell}(w, z) \big)^\beta \text{ for all } \mu, \nu \in \calP(\sfE_\ell),\; w, z \in \sfE_\ell,
              \end{equation}
        \item In the decoder $\Gamma_p$, every Markov kernel $K_h: \calP(\sfE_L) \times \sfE_L \to \calP(\sfE_L)$ is H\"older stable in the context: there exists $C > 0$ and $\beta \in (0,1]$ such that
              \begin{equation}
                  W_1\big(K_{h}[\mu,z], K_{h}[\nu,z] \big) \le C  W_1(\mu, \nu) ^\beta \qquad \text{for all } \mu, \nu \in \calP(\sfE_L),\; z \in \sfE_L.
              \end{equation}
    \end{enumerate}
\end{definition}

\begin{remark} In particular, the decoder $\Gamma_p$ is not required to be continuous with respect to its second argument $z = q_\sfX(x)$.
\end{remark}

\begin{restatable}{theorem}{representationstability}
    \label{thm:representation-1}
    Every stable Markov kernel network $\calN: \calP(\sfG) \times \sfX \to \sfY$ is uniformly discretization stable on $\calS = \calP(\sfG)$.
\end{restatable}

Our main result gives a constructive extension of operators that are uniformly discretization stable on admissible sets of non-collapsing measures to uniformly discretization stable operators on the entire graph space, with the extension being a stable Markov kernel network.

\begin{restatable}{theorem}{representationextension}
    \label{thm:representation-2}
    Let $\calS \subset \calP(\sfG)$ be an admissible set of non-collapsing measures, and $\calN^*: \calS \times \sfX \to \sfY$ be uniformly discretization stable on $\calS$. Then there exists a stable Markov kernel network $\calN: \calP(\sfG) \times \sfX \to \sfY$ such that $\calN|_{\calS \times \sfX} = \calN^*$.
\end{restatable}

The main idea in the proof is to normalize spatial features and calibrate the decoder reads using techniques from \citet{decarvalho2013euclidean}. The proof is in Appendix \ref{sec:proofs}.

\begin{example}[Counterexample when $\calS = \calP(\sfG)$]
    \label{example:counterexample}
We cannot use stable Markov kernel networks to represent all uniformly discretization stable operators without restricting the domain of input measures $\calS$. Let $\sfX = \sfY = [0, 1]$ equipped with the Lebesgue measure $\lambda$, $\sfG=\sfX\times \sfY$ and consider $\calN^*: \calP(\sfG) \times \sfX \to \sfY$ defined by
    \begin{equation}
        \calN^*\left( \mu, x \right) = \sqrt{\int_\sfG \tilde x \; \mu(\dd\tilde x, \dd y)}.
    \end{equation}
    This operator is uniformly discretization stable on the entire domain $\calS = \calP(\sfG).$
It is constant on graph measures, $\calN^*(\gamma_f, x) = 1 / \sqrt{2}$ for all $f \in L^2(\sfX, \sfY)$. However, for Dirac inputs $\delta_{(t,0)}, t \in [0,1]$, the map $t \mapsto \calN^*(\delta_{(t,0)}, x) = \sqrt{t}$ is not Lipschitz. On the other hand, every stable Markov kernel network $\calN$ forces the map $t \mapsto \calN(\delta_{(t,0)}, x)$ to be Lipschitz (the support condition \eqref{eq:markov_kernel_support} forces every read head for Dirac inputs to be $r_h(\delta_z, z) = v_h(z)$, and the stable network becomes a composition of Lipschitz maps).
\end{example}

Example \ref{example:counterexample} illustrates the importance of the non-collapse condition \eqref{eq:admissibility-criterion} on the domain $\calS$ in Theorem \ref{thm:representation-2}. We can match the behavior of $\calN^*$ with a stable Markov kernel network on any reasonable discretized input measure, even very noisy approximations of graph measures, but not its behavior when the spatial marginal collapses. However, the theorem shows that when we restrict the domain of $\calN^*$ to an admissible class $\calS$, we can construct a stable Markov network that matches the behavior of $\calN^*$ exactly on $\calS$, and the resulting network is uniformly discretization stable on all of $\calP(\sfG)$.

We close this section by characterizing the operators represented by stable Markov kernel networks.

\begin{restatable}[Operator characterization of stable Markov kernel networks]{corollary}{operatorcharacterization}
    Let $A: L^2(\sfX, \sfY) \to L^2(\sfX, \sfY)$ be a possibly non-linear operator mapping functions to functions.
    Then there exists $C > 0$, $\beta \in (0,1]$ such that
    \begin{equation}
        \| A[f] - A[g] \|_{L^\infty(\sfX, \R^\dout)}
        \le C W_1(\gamma_f, \gamma_g)^\beta \qquad \text{for all } f, g \in L^2(\sfX, \sfY)
    \end{equation}
    if and only if there exists a stable Markov kernel network $\cal{N}: \calP(\sfG) \times \sfX \to \sfY$ such that
    \begin{equation}
        A[f](x) = \calN(\gamma_f, x) \qquad \text{for all } f \in L^2(\sfX, \sfY) \text{ and $\lambda$-a.e. } x \in \sfX.
    \end{equation}
\end{restatable}

\section{Variational principles for Markov kernel networks}
\label{sec:variational-principles}

Theorems \ref{thm:representation-1} and \ref{thm:representation-2} define the design space of uniformly discretization stable neural networks to be the stable Markov kernel networks, and Definition \ref{def:stable-markov-kernel-network} gives stability properties of the individual components $K_h, v_h, m, q_\sfG, q_\sfX$ in such networks. Exploring new architectures in the design space becomes a question of specifying parameterizations or fixed implementations of these components.
We propose one approach for designing the Markov kernels $K_h$ in~\eqref{eq:read-head} based on variational principles.
We fix
$q_\sfG$, $q_\sfX$, and every $v_h$ and $m$. The proof of Theorem \ref{thm:representation-2} shows this can be done in a manner that does not diminish the expressive power of the network itself.

Our approach is inspired by the variational characterization of softmax from \citet{niculae2017regularized}. For key, query matrices $K_{\mathrm{key}}, Q$ and temperature $\tau > 0$, the minimizer of
\[
\operatorname*{arg\,min}_{\kappa\in\mathcal P(\sfE)}
\left\{
-\int_\sfE
\langle Qz,\,K_{\mathrm{key}}w\rangle
\,\kappa(\dd w)
+
\tau\,\mathrm{KL}(\kappa\Vert\mu)
\right\}
\]
is the classical softmax attention kernel
\[
K[\mu,z](\dd w)
=
\frac{
\exp\!\left({\langle Qz,\,K_{\mathrm{key}}w\rangle}/{\tau}
\right)
}{
\int_\sfE
\exp\!\left(
{\langle Qz,\,K_{\mathrm{key}}w'\rangle}/{\tau}
\right)
\,\mu(\dd w')
}
\,\mu(\dd w).
\]

We consider a general cost $c: \calP(\sfE) \times \sfE \times \sfE \to \R$ and a regularizer $\calR: \calP(\sfE) \times \calP(\sfE) \to [0, +\infty]$. Set
\begin{equation}
    \label{eq:objective-function-J}
    K[\mu, z] \in
    \argmin_{\substack{
        \kappa \in \calP(\sfE) \\
        \supp \kappa \subseteq \supp \mu
    }}
    J_{\mu,z}(\kappa).\qquad J_{\mu,z}(\kappa)
    :=
    \int_\sfE c(\mu, z, w) \kappa(\dd w)+ \calR(\kappa, \mu).
\end{equation}
Here, $c(\mu, z, w)$ is the cost of reading from location $w$ based on the input context $\mu$ and query $z$.
The regularizer $\calR$ penalizes large deviations from $\mu$.
The support condition restricts reads to the context $\mu$, so when a minimizer $K$ exists, it provides legal read behavior for a Markov kernel network consistent with Definition \ref{def:markov-kernel-in-context-map}.

From this point of view, the design space is $(c,\calR)$-space.  We ask: what are the conditions on costs and regularizers that guarantee the associated kernel is sufficiently stable? We can answer this question for a large class of regularizers constructed from $f$-divergences and Wasserstein distances.

\subsection{Regularizers from $f$-divergences}

An $f$-divergence $D_\varphi: \calP(\sfE) \times \calP(\sfE) \to [0, +\infty]$ with generator $\varphi: [0,+\infty) \to \R\cup\{+\infty\}$ is of the form
\begin{equation}
    D_\varphi(\kappa \, \| \, \mu) =
    \begin{cases}
        \int_\sfE \varphi(\frac{\dd\kappa}{\dd\mu})\,\dd\mu, \qquad &\kappa \ll \mu \\
        \infty, \qquad &\kappa \not\ll \mu.
    \end{cases}
\end{equation}
where $\varphi$ is convex and $\varphi(1) = 0$. We further define $\varphi(0) = \lim_{t \to 0}\varphi(t),$ which could be infinite.
This class covers many common divergences, including KL, Hellinger, Jensen-Shannon, and $\chi^2$.
Not all $f$-divergence regularizers lead to Markov kernels with our required stability properties; the next proposition gives sufficient conditions on the $f$-divergence regularizer and cost function for this to happen.

\begin{proposition}[Kernel stability for $f$-divergence regularizers]
    \label{prop:f-divergences}
    Let $c: \calP(\sfE) \times \sfE \times \sfE \to \R$ satisfy
    \begin{equation}
\label{cost-regularity}
\begin{split}
    |c(\mu, z, w)| &\leq C, \\
    |c(\mu, z, w) - c(\mu', z', w')| &\leq L \rho_\sfE(w, w') + L (\rho_\sfE(z, z') + W_1(\mu, \mu'))^\beta,
\end{split}
\end{equation}
for some $C, L > 0$, $\beta \in (0, 1].$
    Let $D_\varphi$ have generator $\varphi: [0,+\infty) \to [0,+\infty]$ that is lower semicontinuous, strictly convex, and finite and $C^1$ on $(0, +\infty)$, with $\varphi(1) = \varphi'(1) = 0$ and $\lim_{r\to+\infty} \varphi'(r) = +\infty$. Further suppose the function
    \begin{equation}
        h(t) = \begin{cases}
            0, &\qquad t \leq \varphi'(0+), \\
            (\varphi')^{-1}(t), &\qquad t > \varphi'(0+)
        \end{cases}
    \end{equation}
    is H\"older stable on $[-2C, 2C]$.
Then
\eqref{eq:objective-function-J} with cost function $c$ and regularizer $\calR = D_\varphi$ has a unique minimizer $K$ satisfying the stability condition \eqref{eq:joint-stability-1}.
Moreover, the kernel takes the form
    \begin{equation}
    \label{eqn:kernel-f-divergence}
        K[\mu,z](\dd w)
        = h\!\left(
            b_{\mu,z}-c(\mu,z,w)
            \right)\mu(\dd w),
            \qquad
            \int_\sfE
            h\!\left(
            b_{\mu,z}-c(\mu,z,w)
            \right)\mu(\dd w)
            =
            1,
    \end{equation}
where \(b_{\mu,z}\in[-C,C]\) is the unique normalizing constant.
\end{proposition}

A proof is in Appendix \ref{app:density-kernel-stability}.
Table \ref{tab:f-divergences} gives a non-exhaustive list of discretization stable Markov kernels generated in this way. We further note that the Pearson $\chi^2$ regularizer corresponds to the construction used in the proof of Theorem~\ref{thm:representation-2}.

\begin{table}[t]
    \centering
    \small
    \setlength{\tabcolsep}{4pt}
    \renewcommand{\arraystretch}{1.3}
    \begin{tabular}{@{}lllp{0.24\linewidth}@{}}
        \toprule
        Divergence & Generator $\varphi(r)$ & Response $h(t)$
        & Attention mechanism \tabularnewline
        \midrule
        KL
        & $r\log r-r+1$
        & $e^t$ & \raggedright Softmax \citep{vaswani2017attention} \tabularnewline
        \addlinespace
        Pearson $\chi^2$
        & $\frac12(r-1)^2$
        & $(1+t)_+$ & \raggedright Sparsemax \citep{martins2016sparsemax} \tabularnewline
        \addlinespace
        \shortstack[l]{Tsallis / power,\\$q>1$}
        & $\displaystyle\frac{r^q-qr+q-1}{q(q-1)}$
        & $[1+(q-1)t]_+^{1/(q-1)}$ & \raggedright Entmax \citep{martins2022sparsecontinuous} \tabularnewline
        \addlinespace
        \shortstack[l]{Shifted entropy,\\$a>0$}
        & $\begin{gathered}(r+a)\log\!\frac{r+a}{1+a}\\{}-r+1\end{gathered}$
        & $[(1+a)e^t-a]_+$ & --- \tabularnewline
        \addlinespace
        \shortstack[l]{KL plus Pearson,\\$a>0$}
        & $\begin{gathered}r\log r-r+1\\{}+\frac a2(r-1)^2\end{gathered}$
        & $a^{-1}W_0(ae^{t+a})$ & --- \tabularnewline
        \addlinespace
        \shortstack[l]{Centered power,\\$p>1$}
        & $\displaystyle\frac1p\lvert r-1\rvert^p$
        & $[1+\operatorname{sgn}(t)\lvert t\rvert^{1/(p-1)}]_+$ & \raggedright Sparsemax for $p=2$ \citep{martins2016sparsemax} \tabularnewline
        \addlinespace
        \shortstack[l]{Quadratic plus\\reverse KL}
        & $\displaystyle\frac{r^2-1-2\log r}{4}$
        & $t+\sqrt{1+t^2}$ & --- \tabularnewline
        \addlinespace
        Jeffreys
        & $(r-1)\log r$
        & $[W_0(e^{1-t})]^{-1}$ & \raggedright $f$-softargmax \citep{roulet2025fdivergences} \tabularnewline
        \bottomrule
    \end{tabular}
    \caption{Examples of $f$-divergences consistent with the hypotheses of Proposition~\ref{prop:f-divergences}. $W_0$ denotes the principal real Lambert $W$ function.
    The associated stable read kernel $K$ is given by Equation \eqref{eqn:kernel-f-divergence}.
    A dash in the last column indicates we are unaware of previous investigations of these attention mechanisms.
    }
    \label{tab:f-divergences}
\end{table}

\subsection{Regularizers from projected Wasserstein distances}

Regularizers constructed from $f$-divergences cover a range of Markov kernel read behavior, but they do not recover the SelfWarp behavior from Flowers. For this, we turn to regularizers built from projected Wasserstein distances. Let $\pi_\sfX: \sfE \to \sfX$ be fixed, continuous and surjective. Define
\begin{equation}
    \label{eq:wasserstein-calR}
    \calR(\kappa, \mu) := \frac12 W_2({\pi_\sfX}_\# \kappa, {\pi_\sfX}_\# \mu)^2.
\end{equation}
The function $\pi_\sfX$ should be interpreted as a projection. The regularizer above penalizes reads in an optimal transport distance on the spatial domain $\sfX$.
\begin{proposition}
    \label{prop:wasserstein-calR-stable}
    Let $\calR$ be given by equation \eqref{eq:wasserstein-calR}. If $(\mu, z, w) \mapsto c(\mu, z, w)$ is continuous in $w$, then there exists a Markov kernel $K$ such that $K[\mu, z]$ satisfies \eqref{eq:objective-function-J} for all $\mu$ and $z$.
\end{proposition}

To recover SelfWarp on continuous graph contexts $\gamma_f$, let
$\tau_h(x,a)=x+g_h(a)\in\sfX$, with periodic wrapping when appropriate.
For $B>\operatorname{diam}(\sfX)$, choose
$$
    c_h(\mu,z,w)=B\rho_\sfX\bigl(\pi_\sfX(w),\tau_h(z)\bigr).
$$
The unique minimizing kernel is
$K_h[\gamma_f,z]=\delta_{(\tau_h(z),f(\tau_h(z)))}$.
With value map $v_h(y,b)=V_hb$, this gives the SelfWarp read
$V_hf(x+g_h(f(x)))$ at $z=(x,f(x))$.
The precise statement and proof are in
Appendix~\ref{app:wasserstein-selfwarp}.

\begin{remark}
    Our final question is: can we construct a single parameterized family of regularizers such that the read behaviors of FNO, transformers, and Flowers are recovered in different parameter regimes? We show this is possible in Appendix~\ref{app:s4-proofs}, which constructs a convex space of regularizers as a linear combination of spatial Wasserstein and $f$-divergences such that the resulting Markov kernel reads are discretization stable. In future work, we aim to explore this space using agents with the goal of adaptively choosing architectures suited to a particular operator learning problem at hand.
\end{remark}

\section{Limitations}
\label{sec:discussion}

The representation theorem presented here does not specify how to choose subclasses of Markov kernel networks suitable for a particular task, nor does it determine the most efficient implementations. For example, although it is possible to implement the GroupNorm operation as part of a Markov kernel layer (see Equation \eqref{eq:flowers-groupnorm} in Section~\ref{sec:concrete-flowers}), it is another question to choose an efficient implementation in code. The practical importance of depth is not reflected in the proof of the representation theorem, which requires only two layers. Additionally, the kernels $K_h$ in the representation might be complicated.

\subsection*{Acknowledgments}

I.D. was partially supported by the European Research Council Consolidator Grant 101232533 (PhaseShift). The research of M.L. was partially supported by European Research Council Advanced Grant 101097198 (PDE-Inverse)  and the Research Council of Finland, grant 359182 (FAME Flagship). MVdH gratefully acknowledges the support of the Department of Energy BES, under grant DE-SC0020345, Oxy and the Simons Foundation under the MATH + X Program. Views and opinions expressed are those of the authors only and do not necessarily reflect those of the European Union or the other funding organizations.

\subsection*{AI use statement}

The main portion of this document was written by humans; there is no LLM-generated text appearing on any page preceding this statement, except for light line-level edits.
The appendices interleave text generated by ChatGPT 6 Astra and humans; all LLM-generated text was reviewed and edited by the authors for clarity and correctness. We also used ChatGPT and Claude for non-writing tasks, such as literature searches and exploring mathematical ideas.

\bibliography{references}

\appendix

\section{Notation and background}
\label{sec:notation-background}

\begingroup
\small
\renewcommand{\arraystretch}{1.12}
\begin{longtable}[l]{@{}>{\raggedright\arraybackslash}p{0.34\linewidth}>{\raggedright\arraybackslash}p{\dimexpr0.66\linewidth-2\tabcolsep\relax}@{}}
    \toprule
    Notation & Description \\
    \midrule
    \endfirsthead
    \toprule
    Notation & Description (continued) \\
    \midrule
    \endhead
    \midrule
    \multicolumn{2}{r@{}}{\footnotesize Continued on next page} \\
    \endfoot
    \bottomrule
    \endlastfoot

    \multicolumn{2}{@{}l}{\textit{Spaces, measures, and norms}} \\*
    \addlinespace[2pt]
    $(\sfX,\rho_\sfX)$ & Compact spatial domain (metric space $\sfX$, metric $\rho_\sfX$). \\
    $\sfY\subset\R^\dout$ & Nonempty compact convex input/output feature range. \\
    $\sfG=\sfX\times\sfY$, $\rho_\sfG$ & Graph space and its product metric. \\
    $\sfE,\sfE',\sfE_\ell$ & Compact embedding spaces; $\ell$ indexes layers. \\
    $\calP(\sfM)$, $\calP_1(\sfM)$ & Radon probability measures; those with finite first moment. \\
    $\mu$, $\nu$, $\kappa$ & Context, comparison measure, and candidate read measure. \\
    $\lambda$ & Reference probability measure with $\supp\lambda=\sfX$. \\
    $\gamma_f=(\identity,f)_\#\lambda$ & Graph measure of $f$. \\
    $\delta_z$, $\supp\mu$ & Dirac probability measure at $z$; support of $\mu$. \\
    $F_\#\mu$ & Pushforward of $\mu$ through $F$. \\
    $\pi_\sfX$, $(\pi_\sfX)_\#\mu$ & Spatial projection and spatial marginal. \\
    $W_1$, $W_2$ & Wasserstein distances of orders one and two. \\
    $L^2(\sfX,\sfY)$ & Square-integrable functions, identified $\lambda$-a.e. \\
    $\|g\|_\infty$, $\|g\|_{L^\infty}$ & Pointwise supremum; $\lambda$-essential supremum. \\

    \addlinespace[6pt]
    \multicolumn{2}{@{}l}{\textit{Operators and Markov kernel networks}} \\*
    \addlinespace[2pt]
    $A$, $A[f]$ & Target function-to-function operator and its output. \\
    $\calN$, $\calN^*$ & Network; target map in the representation theorem. \\
    $\calS\subseteq\calP(\sfG)$ & Input class; admissible when non-collapsing (Definition~\ref{def:non-collapsing-measures}). \\
    $q_\sfG$, $q_\sfX$ & Input-token and output-query encoders. \\
    $L$, $\mu_\ell$ & Number of hidden context layers; context at layer $\ell$. \\
    $H$, $h$, $d_h$ & Head count, head index, and value/read dimension. \\
    $K_h[\mu,z]$ & Read probability measure supported on $\supp\mu$. \\
    $v_h$ & Value map of head $h$. \\
    $r_h(\mu,z)$ & Read vector $\int_\sfE v_h(w)\,K_h[\mu,z](\dd w)$. \\
    $m$ & Pointwise channel mixer. \\
    $\Gamma(\mu,z)$ & In-context token update combining the query and reads. \\
    $B_\Gamma$, $B_\ell$ & Context pushforward $\Gamma(\mu,\cdot)_\#\mu$; layer $\ell$. \\
    $\Gamma_p$ & Decoder in-context map. \\
    $\beta\in(0,1]$ & H\"older exponent in stability bounds. \\

    \addlinespace[6pt]
    \multicolumn{2}{@{}l}{\textit{Variational reads}} \\*
    \addlinespace[2pt]
    $c(\mu,z,w)$ & Cost of reading source token $w$ at query $z$. \\
    $\calR(\kappa,\mu)$, $J_{\mu,z}(\kappa)$ & Regularizer and cost-plus-regularizer objective. \\
    $\varphi$, $D_\varphi$ & Convex divergence generator and its $f$-divergence. \\
    $h(t)$ & Scalar response of the divergence generator. \\
    $b_{\mu,z}$ & Shift normalizing the read density in \eqref{eqn:kernel-f-divergence}. \\
\end{longtable}
\addtocounter{table}{-1}
\endgroup

A Borel probability measure $\mu$ on a metric space $\sfM$ is Radon if
$$
    \mu(A)=\sup\{\mu(K):K\subseteq A\text{ is compact}\}
$$
for every Borel set $A$. Every Borel probability measure on a compact metric space is Radon.

For a Borel map $F:\sfM\to\sfM'$, the pushforward of $\mu$ is
\begin{equation}
    \label{def:pushforward}
    F_\#\mu(B)=\mu(F^{-1}(B)),\qquad B\subseteq\sfM'\text{ Borel}.
\end{equation}
Equivalently, $\int\phi\,\dd(F_\#\mu)=\int\phi\circ F\,\dd\mu$ for every bounded Borel function $\phi$.

\begin{samepage}
\begin{definition}[Wasserstein distances]
    \label{def:wasserstein-1}
    For $1\le p<\infty$, let $\calP_p(\sfM)$ denote the Radon probability measures satisfying $\int\rho_\sfM(z,z_0)^p\,\mu(\dd z)<\infty$ for one, and hence every, $z_0\in\sfM$. For $\mu,\nu\in\calP_p(\sfM)$, define
    \begin{equation}
        W_p(\mu,\nu):=\left(\inf_{\pi\in\Pi(\mu,\nu)}
            \int_{\sfM\times\sfM}\rho_\sfM(z,w)^p\,\pi(\dd z,\dd w)\right)^{1/p},
    \end{equation}
    where $\Pi(\mu,\nu)$ is the set of probability measures on $\sfM\times\sfM$ with marginals $\mu$ and $\nu$.
\end{definition}
\end{samepage}

On a compact metric space, $\calP_p(\sfM)=\calP(\sfM)$ and $W_p$ metrizes weak convergence for every finite $p$ \citep[Chapter~6]{villani2009optimal}. This applies to $\sfG$ and the embedding spaces $\sfE_\ell$ throughout the paper.

We regard $L^2(\sfX,\sfY)$ as equivalence classes modulo equality $\lambda$-almost everywhere. Borel representatives are used when pointwise values are needed. The graph measure is independent of this choice, and $\gamma_f=\gamma_g$ if and only if $f=g$ $\lambda$-almost everywhere. An $L^\infty$ norm is an essential supremum; the supremum in \eqref{eq:DS} is over every $x\in\sfX$.

\subsection{Sequences and empirical measures}
A sequence $z_1,\ldots,z_N$ determines the empirical measure $\mu=N^{-1}\sum_i\delta_{z_i}$. For fixed $N$, this identifies sequences up to permutation. For two equally weighted empirical measures on a metric space,
$$
    W_1\left(\frac1N\sum_i\delta_{z_i},\frac1N\sum_i\delta_{w_i}\right)
    =\min_{\sigma\in S_N}\frac1N\sum_i\rho_\sfM(z_i,w_{\sigma(i)}).
$$
The same measure can arise from different sequence lengths by repeating every token. A map defined on measures must give the same result under such repetitions. The attention formula below has this property.

\subsection{Softmax attention as a context pushforward}
\label{sec:attention-as-pushforward}
Consider tokens $z_1,\ldots,z_N$ in a compact subset $\sfE$ of a Euclidean space. For query, key, value, and output matrices $Q,K_{\mathrm{key}},V,W$, put
$$
    s(z,w)=\frac{\langle Qz,K_{\mathrm{key}}w\rangle}{\sqrt{k}},
    \qquad
    K[\mu,z](\dd w)=\frac{e^{s(z,w)}}{\int_\sfE e^{s(z,u)}\,\mu(\dd u)}\,\mu(\dd w),
$$
where $k$ is the query/key dimension. The value map is $v(w)=Vw$, and the mixer $m(z,r)=z+Wr$ includes a residual connection. For $\mu=N^{-1}\sum_j\delta_{z_j}$, this gives
\begin{equation}
    \Gamma(\mu,z_i)=z_i+W\sum_{j=1}^N
        \frac{e^{s(z_i,z_j)}}{\sum_{q=1}^Ne^{s(z_i,z_q)}}Vz_j,
    \qquad
    B_\Gamma(\mu)=\frac1N\sum_{i=1}^N\delta_{\Gamma(\mu,z_i)}.
\end{equation}
This is the usual single-head softmax attention update, written as $B_\Gamma(\mu)=\Gamma(\mu,\cdot)_\#\mu$. For a weighted context $\sum_j\omega_j\delta_{z_j}$, the attention weights are proportional to $\omega_j e^{s(z_i,z_j)}$. Section~\ref{app:transformers} gives the full block, including multiple heads, normalization, and the feed-forward map.

\section{Proofs}
\label{sec:proofs}

\subsection{Section 3}

\representationstability*

\begin{proof}
    We first derive stability of the in-context maps from the component assumptions in Definition~\ref{def:stable-markov-kernel-network}. Consider a hidden layer or the decoder, with input space $\sfE$, output space $\sfE'$, and $H$ heads. Write $A_h>0$ and $\beta_h\in(0,1]$ for its kernel stability constants, and let $V_h>0$ and $M>0$ be Lipschitz constants for its value maps and mixer, using the sum product metric for the mixer. Set
    $$
        \beta=\min_{1\le h\le H}\beta_h,
        \qquad D=\max\{1,2\operatorname{diam}(\sfE)\}.
    $$
    Since every distance in the kernel bounds is at most $D$, the inequality $t^{\beta_h}\le D^{\beta_h-\beta}t^\beta$ allows us to use the common exponent $\beta$ after enlarging the constants $A_h$.

    For probability measures $\alpha,\alpha'\in\calP(\sfE)$ and any coupling $\pi$ of them, Lipschitz continuity of $v_h$ gives
    $$
        \left\|\int v_h\,\dd\alpha-\int v_h\,\dd\alpha'\right\|
        \le \int\|v_h(u)-v_h(v)\|\,\pi(\dd u,\dd v)
        \le V_h\int\rho_\sfE(u,v)\,\pi(\dd u,\dd v).
    $$
    Taking the infimum over couplings, each read therefore satisfies
    $$
        \|r_h(\mu,w)-r_h(\nu,z)\|
        \le V_h W_1\big(K_h[\mu,w],K_h[\nu,z]\big).
    $$
    For a hidden layer, put $t=W_1(\mu,\nu)+\rho_\sfE(w,z)$. Joint kernel stability and the Lipschitz mixer imply
    \begin{align*}
        \rho_{\sfE'}\big(\Gamma(\mu,w),\Gamma(\nu,z)\big)
        &\le M\left(\rho_\sfE(w,z)+\sum_{h=1}^H V_h A_h t^\beta\right) \\
        &\le M\left(D^{1-\beta}+\sum_{h=1}^H V_h A_h\right)t^\beta.
    \end{align*}
    Here $t\le D$ and $\beta\le1$ imply $t\le D^{1-\beta}t^\beta$. For the decoder, the same argument at a fixed query gives
    $$
        \rho_{\sfE'}\big(\Gamma(\mu,z),\Gamma(\nu,z)\big)
        \le M\left(\sum_{h=1}^H V_h A_h\right)W_1(\mu,\nu)^\beta,
    $$
    uniformly in $z$. Applying these estimates to each layer yields joint stability constants $C_\ell>0$, $\beta_\ell\in(0,1]$ for $\Gamma_\ell$ and context-only stability constants $C_p>0$, $\beta_p\in(0,1]$ for $\Gamma_p$.

    We next show that each context update $B_\ell$ is H\"older continuous in $W_1$. Fix $\eta,\xi\in\calP(\sfE_\ell)$, write $d=W_1(\eta,\xi)$, and choose an optimal coupling $\pi$ of $\eta$ and $\xi$. Such a coupling exists because $\sfE_\ell$ is compact. The pushforward of $\pi$ under
    \[
        (w,z)\longmapsto\big(\Gamma_\ell(\eta,w),\Gamma_\ell(\xi,z)\big)
    \]
    is a coupling of $B_\ell(\eta)$ and $B_\ell(\xi)$. Consequently, joint stability gives
    \begin{align*}
        W_1\big(B_\ell(\eta),B_\ell(\xi)\big)
        &\le C_\ell\int\big(d+\rho_{\sfE_\ell}(w,z)\big)^{\beta_\ell}\,\pi(\dd w,\dd z) \\
        &\le C_\ell\left(d+\int\rho_{\sfE_\ell}(w,z)\,\pi(\dd w,\dd z)\right)^{\beta_\ell} \\
        &=2^{\beta_\ell}C_\ell\,d^{\beta_\ell}.
    \end{align*}
    The second inequality is Jensen's inequality for the concave function $t\mapsto t^{\beta_\ell}$, and the last equality uses optimality of $\pi$.

    Now fix arbitrary inputs $\mu,\nu\in\calP(\sfG)$, and denote their successive encoded contexts by $\mu_\ell,\nu_\ell$. If $Q>0$ is a Lipschitz constant for $q_\sfG$, then
    \[
        W_1(\mu_0,\nu_0)
        =W_1\big({q_\sfG}_\#\mu,{q_\sfG}_\#\nu\big)
        \le Q\,W_1(\mu,\nu).
    \]
    Iterating the context-update estimate therefore yields a constant $A>0$, depending only on the network, such that
    \[
        W_1(\mu_L,\nu_L)\le A\,W_1(\mu,\nu)^\theta,
        \qquad \theta=\prod_{\ell=0}^{L-1}\beta_\ell.
    \]

    At each output location $x$, both evaluations use the same query $q_\sfX(x)$. The decoder's stability bound is uniform over these queries, so
    \begin{align*}
        \sup_{x\in\sfX}|\calN(\mu,x)-\calN(\nu,x)|
        &=\sup_{x\in\sfX}\big|\Gamma_p(\mu_L,q_\sfX(x))-\Gamma_p(\nu_L,q_\sfX(x))\big| \\
        &\le C_p\,W_1(\mu_L,\nu_L)^{\beta_p} \\
        &\le C_p A^{\beta_p}\,W_1(\mu,\nu)^{\beta_p\theta}.
    \end{align*}
    Since $\beta_p\theta\in(0,1]$ and $\mu,\nu$ were arbitrary, this is uniform discretization stability on $\calP(\sfG)$.
\end{proof}

\representationextension*

    \begin{proof}
    We use one context layer to calibrate finitely many spatial values so that they have zero mean and unit total variance on $\calS$. The decoder then recovers the target through weighted averages, followed by a fixed linear readout and projection onto $\sfY$.

    \emph{Step 1: Extend the target in the context variable.}
    Let $C>0$ and $\beta\in(0,1]$ be the discretization-stability constants of $\calN^*$, write $c=\dout$, and choose $M\ge1$ such that $\sfY\subset[-M,M]^c$. Since $\calP(\sfG)$ is a compact metric space, $\calS$ has a countable dense sequence $(\mu^{(i)})_{i\ge1}$. For $j=1,\ldots,c$, define
    \[
        U_j(\mu,x)
        =\operatorname{clip}_{[-M,M]}\!\left(
        \inf_{i\ge1}\left\{\calN_j^*(\mu^{(i)},x)
            +C\,W_1(\mu,\mu^{(i)})^\beta\right\}\right),
        \qquad (\mu,x)\in\calP(\sfG)\times\sfX,
    \]
    where $\operatorname{clip}_{[-M,M]}(t)=\max\{-M,\min\{t,M\}\}$. This is Borel measurable because the infimum is countable. The triangle inequality for the metric $W_1^\beta$, followed by the fact that clipping is $1$-Lipschitz, gives
    \[
        |U_j(\mu,x)-U_j(\nu,x)|\le C\,W_1(\mu,\nu)^\beta.
    \]
    If $\mu\in\calS$, discretization stability bounds every term in the infimum from below by $\calN_j^*(\mu,x)$, while density provides a sequence of terms converging to that value. Thus $U=\calN^*$ on $\calS\times\sfX$. Globally, $|U|\le\sqrt{c}\,M$ and $U$ is H\"older continuous in the context, uniformly in $x$, with constant $C_U=\sqrt{c}\,C$.

    \emph{Step 2: Find finitely many spatial values with uniformly positive total variance.}
    The set $\calT=\overline{\calS}$ is compact and remains non-collapsing: indeed,
    \[
        \mu\longmapsto\inf_{a\in\sfX}W_1\big((\pi_\sfX)_\#\mu,\delta_a\big)
    \]
    is $1$-Lipschitz. Consequently, every $\mu\in\calT$ has a non-Dirac spatial marginal. For each such measure, some Lipschitz function $h:\sfX\to[0,1]$ has positive variance; one may take a rescaled distance from a point in the marginal's support. That variance remains positive in a neighborhood of $\mu$. A finite subcover of $\calT$ therefore provides Lipschitz functions $h_1,\ldots,h_n:\sfX\to[0,1]$ whose variances have a strictly positive sum throughout $\calT$.

    Write $\phi_k=h_k\circ\pi_\sfX$ and, for every $\mu\in\calP(\sfG)$, set
    \[
        a_k(\mu)=\int_{\sfG}\phi_k\,\dd\mu,
        \qquad b_k(\mu)=\int_{\sfG}\phi_k^2\,\dd\mu-a_k(\mu)^2,
        \qquad V(\mu)=\sum_{k=1}^n b_k(\mu).
    \]
    These functions are globally Lipschitz in $W_1$, since the $\phi_k$ are bounded and Lipschitz. By compactness there is $v_*>0$ such that $V(\mu)\ge v_*$ on $\calT$. Define
    \[
        D(\mu)=\max\{V(\mu),v_*\}.
    \]

    \emph{Step 3: Calibrate the values in a context layer.}
    Take $L=1$ and set
    \[
        R=v_*^{-1/2},\qquad
        \sfE_0=\sfG,\qquad
        \sfE_1=\sfG\times[-R,R]^n.
    \]
    Equip $\sfE_1$ with the sum of the graph metric and the Euclidean metric on the new coordinates. Choose $y_*\in\sfY$ and use the $1$-Lipschitz encoders
    \[
        q_\sfG=\identity_\sfG,
        \qquad q_\sfX(x)=\big((x,y_*),0\big).
    \]
    The context layer retains each original token and appends its calibrated spatial values:
    \[
        \Gamma_0(\mu,g)
        =\left(g,
            \left(\frac{\phi_k(g)-a_k(\mu)}{\sqrt{D(\mu)}}\right)_{k=1}^n
        \right).
    \]
    This is a Markov kernel in-context map with one head: use the uniform kernel $\widehat K[\mu,g]=\mu$ and the value map
    \[
        \widehat v(w)=\big(\phi_k(w),\phi_k(w)^2\big)_{k=1}^n.
    \]
    Its read supplies all first and second moments needed to compute $a_k$ and $D$, so the mixer can apply the displayed formula using only the token and this read. To define the mixer on arbitrary read vectors, first clip every moment coordinate to $[0,1]$. Actual reads are unchanged, and the resulting feature coordinates always lie in $[-R,R]$.

    The uniform kernel $\widehat K$ is jointly $1$-Lipschitz, and $\widehat v$ is Lipschitz because the $\phi_k$ are bounded and Lipschitz. The mixer is globally Lipschitz as well: after clipping, its moment coordinates lie in $[0,1]$, squaring is Lipschitz on this interval, and $t\mapsto\max\{t,v_*\}^{-1/2}$ is bounded and Lipschitz. Its feature numerators are bounded and Lipschitz in the token and clipped moments. Thus the hidden layer satisfies the component assumptions of Definition~\ref{def:stable-markov-kernel-network} on its full domain.

    Write $\pi:\sfE_1\to\sfG$ for projection onto the retained token and $\psi_k:\sfE_1\to[-R,R]$ for the $k$th new coordinate. For any $\eta\in\calP(\sfE_1)$, define
    \[
        \bar a_k(\eta)=\int_{\sfE_1}\psi_k\,\dd\eta,
        \qquad
        T(\eta)=\sum_{k=1}^n
            \left(\int_{\sfE_1}\psi_k^2\,\dd\eta-\bar a_k(\eta)^2\right).
    \]
    For the updated context $\eta=B_0(\mu)=\Gamma_0(\mu,\cdot)_\#\mu$, direct integration gives
    \[
        \pi_\#\eta=\mu,\qquad
        \bar a_k(\eta)=0,\qquad
        T(\eta)=\frac{V(\mu)}{D(\mu)}.
    \]
    Thus the new coordinates have zero mean for every input, and their total variance is one whenever $\mu\in\calS$. This completes the calibration.

    \emph{Step 4: Decode with weighted averages and a fixed readout.}
    For arbitrary $\eta\in\calP(\sfE_1)$ and $z\in\sfE_1$, set
    \[
        \theta_j(\eta,z)
        =\frac{U_j(\pi_\#\eta,\pi_\sfX(\pi z))}{4MR},
        \qquad j=1,\ldots,c.
    \]
    Use $cn$ decoder heads, indexed by $(j,k)$, with
    \[
        K_{jk}[\eta,z](\dd w)
        =\big[1+\theta_j(\eta,z)(\psi_k(w)-\bar a_k(\eta))\big]\,\eta(\dd w),
        \qquad v_{jk}(w)=\psi_k(w).
    \]
    Since $|\theta_j|\le1/(4R)$ and $|\psi_k-\bar a_k|\le2R$, the densities lie in $[1/2,3/2]$. Subtracting $\bar a_k(\eta)$ makes them integrate to one even for contexts outside the image of $B_0$. These are Borel probability kernels supported on $\supp\eta$, and the value maps are bounded and Lipschitz. Their reads satisfy
    \[
        r_{jk}(\eta,z)=\bar a_k(\eta)
            +\theta_j(\eta,z)
                \left(\int\psi_k^2\,\dd\eta-\bar a_k(\eta)^2\right).
    \]
    Let $\Pi_\sfY:\mathbb R^c\to\sfY$ be nearest-point projection, which is $1$-Lipschitz because $\sfY$ is non-empty, compact, and convex. The decoder mixer is simply
    \[
        m_p\big(z,(r_{jk})_{j,k}\big)
        =\Pi_\sfY\!\left(
            4MR\left(\sum_{k=1}^n r_{jk}\right)_{j=1}^c
        \right).
    \]
    This mixer is globally Lipschitz, being a fixed linear map followed by the $1$-Lipschitz projection. There is no reference head or variance normalization in this mixer. Indeed, writing $A(\eta)=\sum_k\bar a_k(\eta)$ and $\mathbf1_c=(1,\ldots,1)\in\mathbb R^c$, the read identities give
    \[
        \Gamma_p(\eta,z)
        =\Pi_\sfY\!\left(
            4MR\,A(\eta)\mathbf1_c
            +T(\eta)\,U(\pi_\#\eta,\pi_\sfX(\pi z))
        \right).
    \]
    It remains to verify context-only stability of the decoder kernels themselves. Fix $j,k$ and $z$, and write
    $$
        F_{\eta,z}(w)=1+\theta_j(\eta,z)(\psi_k(w)-\bar a_k(\eta)),
        \qquad d=W_1(\eta,\xi),\qquad \Delta=\operatorname{diam}(\sfE_1).
    $$
    Since $\pi$ and $\psi_k$ are $1$-Lipschitz, Step 1 and the averaging estimate in the proof of Theorem~\ref{thm:representation-1} give
    $$
        |\theta_j(\eta,z)-\theta_j(\xi,z)|\le\frac{C}{4MR}d^\beta,
        \qquad |\bar a_k(\eta)-\bar a_k(\xi)|\le d.
    $$
    Together with $|\theta_j|\le1/(4R)$ and $|\psi_k-\bar a_k|\le2R$, these imply
    $$
        \|F_{\eta,z}\|_\infty\le\frac32,
        \qquad \operatorname{Lip}_w(F_{\eta,z})\le\frac1{4R},
        \qquad \|F_{\eta,z}-F_{\xi,z}\|_\infty
        \le\frac{C}{2M}d^\beta+\frac{d}{4R}.
    $$
    For a $1$-Lipschitz test function $f$ normalized to vanish at a fixed point of $\sfE_1$, we have $\|f\|_\infty\le\Delta$ and $\operatorname{Lip}(fF_{\eta,z})\le3/2+\Delta/(4R)$. Therefore
    \begin{align*}
        \left|\int f\,\dd K_{jk}[\eta,z]-\int f\,\dd K_{jk}[\xi,z]\right|
        &\le\left|\int fF_{\eta,z}\,\dd(\eta-\xi)\right|
            +\left|\int f(F_{\eta,z}-F_{\xi,z})\,\dd\xi\right| \\
        &\le\left(\frac32+\frac{\Delta}{2R}\right)d
            +\frac{\Delta C}{2M}d^\beta.
    \end{align*}
    Kantorovich--Rubinstein duality, followed by $d\le\Delta$ and $\beta\le1$, now gives
    $$
        W_1\big(K_{jk}[\eta,z],K_{jk}[\xi,z]\big)\le C_K W_1(\eta,\xi)^\beta
    $$
    for a constant $C_K$ independent of $z$. This holds for all $\eta,\xi\in\calP(\sfE_1)$, including contexts outside the image of $B_0$. All components therefore satisfy Definition~\ref{def:stable-markov-kernel-network}, and Theorem~\ref{thm:representation-1} gives uniform discretization stability of the assembled network on its full domain.

    Finally, on every context $\eta=B_0(\mu)$ the term $A(\eta)$ vanishes, so
    \[
        \calN(\mu,x)
        =\Gamma_p\big(B_0(\mu),q_\sfX(x)\big)
        =\Pi_\sfY\!\left(\frac{V(\mu)}{D(\mu)}\,U(\mu,x)\right).
    \]
    For $\mu\in\calS$, the ratio $V/D$ is one and $U(\mu,x)=\calN^*(\mu,x)\in\sfY$, proving the required equality. Only the decoder kernels use the query dependence of $U$; no spatial regularity of the target is needed.
    \end{proof}

\operatorcharacterization*

    \begin{proof}
    Suppose first that the displayed H\"older estimate holds, and set
    \[
        \calS=\{\gamma_f:f\in L^2(\sfX,\sfY)\}.
    \]
    This set is non-empty because $\sfY$ is non-empty. Every $\mu\in\calS$ has spatial marginal $(\pi_\sfX)_\#\mu=\lambda$. Since $\lambda$ has full support and $\sfX$ contains at least two points, $\lambda$ is not a Dirac measure. The function
    \[
        a\longmapsto W_1(\lambda,\delta_a)
        =\int_\sfX\rho_\sfX(x,a)\,\lambda(\dd x)
    \]
    is continuous and strictly positive on the compact space $\sfX$. Its minimum $c_\lambda$ is therefore positive, and
    \[
        \inf_{\mu\in\calS}\inf_{a\in\sfX}
        W_1\big((\pi_\sfX)_\#\mu,\delta_a\big)=c_\lambda>0.
    \]
    Taking $\epsilon=c_\lambda/2$ shows that $\calS$ is an admissible set of non-collapsing measures.

    Since $\sfY$ is compact, every output of $A$ belongs to $L^\infty(\sfX,\mathbb R^{\dout})$. Define
    \[
        Q:\calS\to L^\infty(\sfX,\mathbb R^{\dout}),
        \qquad Q(\gamma_f)=A[f].
    \]
    This is well-defined because equality of graph measures implies equality of the input $L^2$ classes. The assumed estimate states that
    \[
        \|Q(\mu)-Q(\nu)\|_{L^\infty}
        \le C\,W_1(\mu,\nu)^\beta
        \qquad (\mu,\nu\in\calS).
    \]
    We next choose representatives of these output classes so that this estimate holds pointwise and the resulting evaluation map is jointly Borel measurable.

    As a subspace of the compact metric space $(\calP(\sfG),W_1)$, the set $\calS$ is separable. Choose a dense sequence $(\mu_i)_{i\ge1}$ in $\calS$ and Borel representatives $u_i:\sfX\to\sfY$ of $Q(\mu_i)$. The set
    \[
        E=\bigcup_{i,j\ge1}
        \left\{x\in\sfX:
            |u_i(x)-u_j(x)|>C\,W_1(\mu_i,\mu_j)^\beta
        \right\}
    \]
    is Borel and $\lambda$-null. Fix $y_*\in\sfY$ and redefine every $u_i$ to equal $y_*$ on $E$. These remain Borel representatives of the same classes and now satisfy
    \[
        |u_i(x)-u_j(x)|\le C\,W_1(\mu_i,\mu_j)^\beta
        \qquad\text{for all }i,j\ge1,\ x\in\sfX.
    \]
    For $k\ge1$ and $\mu\in\calS$, let
    \[
        i_k(\mu)=\min\{i\ge1:W_1(\mu,\mu_i)<2^{-k}\}.
    \]
    Density makes this index well-defined, and it is Borel measurable because each set on which it equals a given index is a relative open ball with finitely many earlier relative open balls removed. Moreover,
    \[
        \sup_{x\in\sfX}
        |u_{i_k(\mu)}(x)-u_{i_\ell(\mu)}(x)|
        \le C(2^{-k}+2^{-\ell})^\beta.
    \]
    Thus the limit
    \[
        \calN^*(\mu,x)=\lim_{k\to\infty}u_{i_k(\mu)}(x)
    \]
    exists uniformly in $x$ and takes values in the closed set $\sfY$. Each approximating map is jointly Borel measurable, so $\calN^*$ is jointly Borel measurable as well. Passing to the limit in the pairwise pointwise bounds gives
    \[
        |\calN^*(\mu,x)-\calN^*(\nu,x)|
        \le C\,W_1(\mu,\nu)^\beta
        \qquad (\mu,\nu\in\calS,\ x\in\sfX).
    \]
    Finally, $Q(\mu_{i_k(\mu)})\to Q(\mu)$ in $L^\infty$, while their chosen representatives converge uniformly to $\calN^*(\mu,\cdot)$. Hence $\calN^*(\mu,\cdot)$ represents $Q(\mu)$.

    The map $\calN^*$ is therefore uniformly discretization stable on $\calS$. Theorem \ref{thm:representation-2} provides a stable Markov kernel network $\calN$ on $\calP(\sfG)\times\sfX$ extending $\calN^*$. For every $f\in L^2(\sfX,\sfY)$, we consequently have $\calN(\gamma_f,x)=A[f](x)$ for $\lambda$-almost every $x$.

    Conversely, suppose such a stable Markov kernel network $\calN$ exists. By Theorem \ref{thm:representation-1}, there are constants $C>0$ and $\beta\in(0,1]$ such that
    \begin{align*}
        \|A[f]-A[g]\|_{L^\infty}
        &=\|\calN(\gamma_f,\cdot)-\calN(\gamma_g,\cdot)\|_{L^\infty}\\
        &\le\sup_{x\in\sfX}|\calN(\gamma_f,x)-\calN(\gamma_g,x)|\\
        &\le C\,W_1(\gamma_f,\gamma_g)^\beta
    \end{align*}
    for all $f,g\in L^2(\sfX,\sfY)$, as required.
    \end{proof}

\subsection{Section 4.1: stable variational reads}
\label{app:density-kernel-stability}

Regularized attention is a familiar instance of \eqref{eq:objective-function-J} \citep{niculae2017regularized,martins2022sparsecontinuous}. A broader family uses $f$-divergences. We denote the convex generator by $\varphi$ to distinguish it from an input field. For a superlinear generator, its divergence and the corresponding regularizer are
\begin{equation}
    \label{eq:f-divergence-regularizer}
    D_\varphi(\kappa\Vert\mu)
    =\begin{cases}
        \displaystyle\int_\sfE\varphi(p)\,\dd\mu,&\kappa=p\mu,\\
        +\infty,&\kappa\not\ll\mu,
    \end{cases}
    \qquad \calR_{\varphi,\tau}=\tau D_\varphi,\quad\tau>0.
\end{equation}
The integral has its extended meaning. The optimizer is obtained by applying the inverse derivative of $\varphi$ to a shifted negative cost \citep{terjek2021moreau}. Quantitative regularity of this scalar response gives a stability certificate even when the reference context changes.

\begin{proposition}[Stable divergence-regularized kernels]
    \label{prop:stable-variational-kernels}
    Let $\sfE$ be compact and $\tau>0$. Let $\varphi:[0,\infty)\to[0,\infty]$ be lower semicontinuous, strictly convex, finite and $C^1$ on $(0,\infty)$, with $\varphi(1)=\varphi'(1)=0$ and $\lim_{r\to\infty}\varphi'(r)=+\infty$. Define its scalar response by
    \begin{equation}
        \label{eq:divergence-response}
        h(u)=\begin{cases}
            0,&u\le\varphi'(0+),\\
            (\varphi')^{-1}(u),&u>\varphi'(0+),
        \end{cases}
    \end{equation}
    where the first case is absent if $\varphi'(0+)=-\infty$. Suppose the Borel cost $c$ satisfies, for some $M,L,H\ge0$ and $\beta\in(0,1]$,
    \begin{equation}
        \label{eq:stable-variational-costs}
        \begin{aligned}
            |c(\mu,z,w)|&\le M,\\
            |c(\mu,z,w)-c(\nu,z',w')|
            &\le L\rho_\sfE(w,w')
            +H\bigl(W_1(\mu,\nu)+\rho_\sfE(z,z')\bigr)^\beta.
        \end{aligned}
    \end{equation}
    Assume that $h$ is $\gamma$-H\"older on $I_M=[-2M/\tau,2M/\tau]$, for some $\gamma\in(0,1]$. Then \eqref{eq:objective-function-J} with $\calR=\calR_{\varphi,\tau}$ has a unique minimizer, defining a supported Borel Markov kernel
    \begin{equation}
        \label{eq:general-variational-density}
        K[\mu,z](\dd w)=h\bigl(b_{\mu,z}-c(\mu,z,w)/\tau\bigr)\,\mu(\dd w),
    \end{equation}
    where $b_{\mu,z}\in[-M/\tau,M/\tau]$ is the unique scalar making the density integrate to one. This kernel satisfies
    \begin{equation}
        \label{eq:variational-kernel-stability}
        W_1\bigl(K[\mu,z],K[\nu,z']\bigr)
        \le C\bigl(W_1(\mu,\nu)+\rho_\sfE(z,z')\bigr)^{\beta\gamma}.
    \end{equation}
    Here $C$ depends only on the displayed bounds, $\operatorname{diam}\sfE$, $h(2M/\tau)$, and the H\"older constant of $h$ on $I_M$. If \eqref{eq:stable-variational-costs} is assumed only for $z'=z$, uniformly in $z$, then the conclusion holds for $z'=z$ with the same uniformity.
\end{proposition}

The normalization $\varphi'(1)=0$ is harmless: subtracting a multiple of $p-1$ does not change the divergence between probability measures. The response condition is a quantitative convexity requirement. For example, a lower bound $\varphi''(p)\ge m_B>0$ for $0<p\le B$, for every finite $B$, suffices for $\gamma=1$, without requiring global strong convexity.

Proposition~\ref{prop:f-divergences} is the case $\tau=1$, $M=C$, and $H=L$ of this result. If the scalar response has exponent $\gamma$, the resulting kernel has exponent $\beta\gamma$.

\begin{proof}
Write $s=(\mu,z)$, $t=(\nu,z')$, $c_s(w)=c(\mu,z,w)$, and
$$
    D=\operatorname{diam}\sfE,\qquad
    a=W_1(\mu,\nu),\qquad
    \delta=a+\rho_\sfE(z,z'),\qquad
    \varepsilon=La+H\delta^\beta.
$$
Let $A$ be a $\gamma$-H\"older constant of $h$ on $I_M$, and set $B=h(2M/\tau)$.

\emph{Normalization and optimality.}
The assumptions on $\varphi$ imply that $h$ is continuous and nondecreasing, strictly increasing wherever positive, with $h(0)=1$, $\lim_{u\to-\infty}h(u)=0$, and $\lim_{u\to\infty}h(u)=\infty$. Thus
$$
    F_s(b):=\int h(b-c_s/\tau)\,\dd\mu
$$
is continuous, goes from zero to infinity, and is strictly increasing wherever positive. It has a unique root $F_s(b_s)=1$. Since $|c_s|\le M$ and $h(0)=1$,
$$
    -M/\tau\le b_s\le M/\tau,
    \qquad
    0\le p_s(w):=h(b_s-c_s(w)/\tau)\le B.
$$
The candidate $p_s\mu$ has finite objective. If $\varphi(0)<\infty$, the generator is bounded on $[0,B]$; if $\varphi(0)=\infty$, then $\varphi'(0+)=-\infty$ and $p_s\ge h(-2M/\tau)>0$, so it is evaluated on a compact subinterval of $(0,\infty)$.

Extend $\varphi$ by $+\infty$ to negative arguments. By the definition of $h$, the scalar $b_s-c_s(w)/\tau$ belongs to $\partial\varphi(p_s(w))$, including when $p_s(w)=0$. For any probability density $q$ of finite divergence, the supporting-line inequality gives
$$
    \varphi(q)-\varphi(p_s)
    \ge (b_s-c_s/\tau)(q-p_s).
$$
Integrating, using $\int(q-p_s)\,\dd\mu=0$, and adding the linear cost proves $J_s(q\mu)\ge J_s(p_s\mu)$. Strict convexity forces $q=p_s$ almost everywhere in the equality case. All other measures have infinite objective, since $c_s$ is bounded. This proves existence and uniqueness, and absolute continuity gives the support constraint. For fixed $b$, $s\mapsto F_s(b)$ is Borel, and
$$
    \{s:b_s<b\}=\{s:F_s(b)>1\}.
$$
Thus the normalizer is Borel, and parameter-dependent integration proves the Markov property.

\emph{Stability without estimating the normalizer.}
Let $\pi$ be an optimal coupling of $\mu,\nu$. The cost assumptions imply
$$
    \int|c_s(w)-c_t(v)|\,\pi(\dd w,\dd v)\le\varepsilon.
$$
On this coupling define $q_0(v)=h(b_s-c_t(v)/\tau)$. Since $h$ is nondecreasing, $q_0-p_t$ has the same sign everywhere. Both $p_s$ and $p_t$ integrate to one against $\pi$, so
$$
    \int|q_0-p_t|\,\dd\pi
    =\left|\int q_0\,\dd\pi-1\right|
    \le\int|q_0-p_s|\,\dd\pi.
$$
Every argument of $h$ lies in $I_M$. The triangle inequality, the response bound, and concavity of $r\mapsto r^\gamma$ therefore give
\begin{equation}
    \label{eq:divergence-density-stability}
    \begin{aligned}
        \int|p_s(w)-p_t(v)|\,\dd\pi
        &\le2A\tau^{-\gamma}\int|c_s(w)-c_t(v)|^\gamma\,\dd\pi\\
        &\le2A(\varepsilon/\tau)^\gamma.
    \end{aligned}
\end{equation}
For any $1$-Lipschitz test function $v$, subtract the midpoint of its range so that $|v|\le D/2$. This does not change its integral against a difference of probability measures. Since $p_s\le B$,
$$
    \left|\int v\,p_s\,\dd\mu-\int v\,p_t\,\dd\nu\right|
    \le Ba+\frac D2\int|p_s(w)-p_t(v')|\,\pi(\dd w,\dd v').
$$
Kantorovich--Rubinstein duality yields the quantitative bound
\begin{equation}
    \label{eq:f-divergence-quantitative-stability}
    W_1(K_s,K_t)\le Ba+DA\tau^{-\gamma}(La+H\delta^\beta)^\gamma.
\end{equation}
If $D=0$, all the measures coincide. Otherwise $a\le\delta\le2D$, so boundedness converts the right-hand side into $C\delta^{\beta\gamma}$. Taking $z'=z$ throughout proves the fixed-query version under only its stated cost hypotheses.
\end{proof}

\subsubsection{KL, Pearson, and Tsallis examples}
\label{app:divergence-examples}

\begin{corollary}[KL and Pearson reads]
    \label{cor:KL-Pearson-kernels}
    Under the cost assumptions of Proposition~\ref{prop:stable-variational-kernels}, the regularizers
    \begin{equation}
        \label{eq:density-regularizers}
        \calR_{\mathrm{KL},\tau}(\kappa,\mu)=\tau\operatorname{KL}(\kappa\Vert\mu),
        \qquad
        \calR_{\mathrm{Pearson},\tau}(\kappa,\mu)=\frac\tau2\chi^2(\kappa\Vert\mu)
    \end{equation}
    yield unique kernels with densities
    \begin{equation}
        \label{eq:variational-density-formulas}
        \begin{aligned}
            p^{\mathrm{KL}}_{\mu,z}(w)
            &=\frac{e^{-c(\mu,z,w)/\tau}}
            {\int_\sfE e^{-c(\mu,z,v)/\tau}\,\mu(\dd v)},\\
            p^{\mathrm{Pearson}}_{\mu,z}(w)
            &=\bigl(b^{\mathrm{Pearson}}_{\mu,z}-c(\mu,z,w)/\tau\bigr)_+.
        \end{aligned}
    \end{equation}
    where $b^{\mathrm{Pearson}}_{\mu,z}$ normalizes the second density. Both satisfy \eqref{eq:variational-kernel-stability} with exponent $\beta$.
\end{corollary}

Indeed, $\varphi_{\mathrm{KL}}(p)=p\log p-p+1$ gives $h(u)=e^u$, while $\varphi_{\mathrm{Pearson}}(p)=(p-1)^2/2$ gives $h(u)=(1+u)_+$. Both responses are Lipschitz on bounded intervals. These are the Gibbs and positive-part transformations underlying softmax and sparsemax \citep{martins2016sparsemax,martins2022sparsecontinuous}. Here $\chi^2(\kappa\Vert\mu)=\int(p-1)^2\,\dd\mu$ for $\kappa=p\mu$, and is infinite otherwise; the factor $1/2$ belongs to the regularizer.

The generators in Corollary~\ref{cor:KL-Pearson-kernels} have the claimed scalar responses, both locally Lipschitz. For KL, normalization gives $e^{b_s}=1/\int e^{-c_s/\tau}\,\dd\mu$. For Pearson, set $b_s^{\mathrm{Pearson}}=1+b_s$ to obtain the stated positive-part density. In the notation of the proof, $B_{\mathrm{KL}}=e^{2M/\tau}$ and $A_{\mathrm{KL}}=B_{\mathrm{KL}}$, while $B_{\mathrm{Pearson}}=1+2M/\tau$ and $A_{\mathrm{Pearson}}=1$. Equation~\eqref{eq:f-divergence-quantitative-stability} therefore gives
\begin{align}
    W_1(K_s^{\mathrm{KL}},K_t^{\mathrm{KL}})
    &\le B_{\mathrm{KL}}a+\frac{DB_{\mathrm{KL}}}\tau\varepsilon,
    \label{eq:KL-quantitative-stability}\\
    W_1(K_s^{\mathrm{Pearson}},K_t^{\mathrm{Pearson}})
    &\le B_{\mathrm{Pearson}}a+\frac D\tau\varepsilon.
    \label{eq:Pearson-quantitative-stability}
\end{align}
No lower bound on the density or on positive empirical atom masses is used.

More generally, the power, or Tsallis, generators and their responses are
\begin{equation}
    \label{eq:Tsallis-variational-response}
    \varphi_q(p)=\frac{p^q-q p+q-1}{q(q-1)},
    \qquad
    h_q(u)=\bigl[1+(q-1)u\bigr]_+^{1/(q-1)},
    \quad q>1.
\end{equation}
The density formula is the continuous entmax transformation \citep[Proposition~8]{martins2022sparsecontinuous}. For $1<q\le2$ the response is locally Lipschitz, so the kernel retains exponent $\beta$. For $q>2$, it is $1/(q-1)$-H\"older, giving exponent $\beta/(q-1)$. These are sufficient exponents, not claims of optimality. The limit $q\to1$ gives KL, and $q=2$ gives our Pearson normalization.

A direct sufficient curvature condition for $\gamma=1$ is $\varphi''(p)\ge m_B>0$ for $0<p\le B$. The mean value theorem then bounds the inverse derivative by $1/m_B$; the same bound holds across a finite zero threshold by taking a limit from positive densities. If $h(I_M)$ is bounded away from zero, the curvature bound is needed only on that smaller density interval. In particular, global strong convexity is unnecessary for KL.

\paragraph{The remaining responses in Table~\ref{tab:f-divergences}.}
Differentiating the shifted-entropy and KL-plus-Pearson generators gives, respectively, $\log((r+a)/(1+a))$ and $\log r+a(r-1)$. Inverting these derivatives, with truncation at zero when necessary, gives the displayed responses. The centered-power derivative is $\operatorname{sgn}(r-1)|r-1|^{p-1}$, so its response is locally H\"older with exponent $\min\{1,1/(p-1)\}$. The derivatives of the quadratic-plus-reverse-KL and Jeffreys generators are $(r-r^{-1})/2$ and $\log r+1-r^{-1}$, yielding the stated formulas. These last two generators have value $+\infty$ at zero, as allowed in the proposition. Apart from the two power families, all responses in the table are Lipschitz on bounded intervals: their generators have strictly positive continuous second derivative on the positive density range, and the shifted-entropy response is also Lipschitz across its zero threshold. All listed generators are normalized, strictly convex, and superlinear.

\paragraph{Why arbitrary divergences are not covered.}
Even smooth strict convexity does not prevent a support discontinuity when the generator has finite asymptotic slope. Consider reverse KL, with $\varphi(p)=-\log p+p-1$, on $\sfE=\{0,1\}$, and take
$$
    \mu_e=(1-e)\delta_0+e\delta_1,\qquad c(w)=-Cw,\qquad C>\tau.
$$
For $0<e<1$, a row with mass $x$ at $1$ has objective $-Cx+\tau\operatorname{KL}(\mu_e\Vert\kappa)$. Its unique minimizer satisfies
$$
    Cx(1-x)=\tau(x-e),\qquad
    x=\frac{C-\tau+\sqrt{(C-\tau)^2+4C\tau e}}{2C}
      \longrightarrow1-\tau/C>0.
$$
At $e=0$, the support constraint forces the row $\delta_0$. Thus no context-continuous selection exists for this bounded Lipschitz cost. Here $\varphi'(p)\to1$, violating the proposition's superlinearity assumption.

Nor do superlinearity and strict convexity alone ensure a H\"older rate. For example, choose a strictly increasing continuous derivative with
$$
    \varphi'(1+r)=\operatorname{sgn}(r)e^{-1/|r|}
    \quad(0<|r|\le1/2),\qquad \varphi'(1)=0,
$$
and extend it strictly increasingly to $(0,\infty)$ with limit $+\infty$ at infinity; integrating from $1$, with the lower-semicontinuous extension at zero, gives a normalized, strictly convex, superlinear generator. On $\sfE=[0,1]$, fix $\mu=(\delta_0+\delta_1)/2$ and take the bounded Lipschitz cost $c(z,w)=\tau\min\{z,e^{-3}\}(2w-1)$. For $0<z\le e^{-3}$ the unique optimal densities at $0,1$ are $1+r(z),1-r(z)$, where $r(z)=1/\log(1/z)$. Hence $W_1(K[\mu,z],K[\mu,0])=r(z)/2$, which obeys no positive H\"older exponent at zero. This explains the quantitative response assumption.

\paragraph{Empirical normalization.}
For $\mu=\sum_i m_i\delta_{w_i}$ with $m_i>0$, write $k_i=m_ip_i$ and $c_i=c(\mu,z,w_i)$. Pearson minimization is exactly
$$
    \min_{k\in\Delta_n}
    \left\{\sum_i c_i k_i+\frac\tau2\sum_i\frac{(k_i-m_i)^2}{m_i}\right\}.
$$
For $m_i=1/n$, the quadratic term differs from $(\tau n/2)\sum_i k_i^2$ by a constant on $\Delta_n$. Completing the square gives the projection of $-c/(\tau n)$. Both quantitative estimates above are independent of the smallest positive $m_i$.

\subsubsection{The fixed Pearson regularizer in the representation theorem}
\label{app:variational-representation}
Use the notation of the proof of Theorem~\ref{thm:representation-2}. The calibration head $\widehat K[\mu,g]=\mu$ is the unique optimizer for $c=0$ and $\calR=\tfrac12\chi^2$. For the decoder, $\psi_k$ is $1$-Lipschitz, $|\psi_k|\le R$, and
$$
    \theta_j(\eta,z)=\frac{U_j(\pi_\#\eta,\pi_\sfX(\pi z))}{4MR},
    \qquad |\theta_j|\le\frac1{4R}.
$$
The decoder costs and corresponding densities are
\begin{equation}
    \label{eq:variational-representation-cost}
    c_{jk}(\eta,z,w)=-\theta_j(\eta,z)\psi_k(w),
\end{equation}
\begin{equation}
    \label{eq:variational-representation-density}
    p^*_{jk}(\eta,z,w)
    =1+\theta_j(\eta,z)\left(\psi_k(w)-\int\psi_k\,\dd\eta\right)
    \in[1/2,3/2].
\end{equation}
Here $|U_j|\le M$ and $U_j$ is $\beta$-H\"older in its context, uniformly at every fixed spatial query. Since $\pi$ is $1$-Lipschitz, the costs in \eqref{eq:variational-representation-cost} obey
$$
    |c_{jk}|\le\tfrac14,\qquad
    \operatorname{Lip}_w(c_{jk})\le\frac1{4R},\qquad
    \sup_w|c_{jk}(\eta,z,w)-c_{jk}(\xi,z,w)|
    \le\frac{C}{4M}W_1(\eta,\xi)^\beta.
$$
They are Borel and satisfy the fixed-query hypotheses globally. The density $p^*_{jk}$ in \eqref{eq:variational-representation-density} integrates to one and lies in $[1/2,3/2]$. Taking $b^{\mathrm{Pearson}}=1-\theta_j\int\psi_k\,\dd\eta$ in \eqref{eq:variational-density-formulas} gives precisely this density. Alternatively, for every probability density $q\in L^2(\eta)$,
$$
    J_{\eta,z}(q\eta)-J_{\eta,z}(p^*_{jk}\eta)
    =\tfrac12\int(q-p^*_{jk})^2\,\dd\eta,
$$
because $c_{jk}+p^*_{jk}-1=-\theta_j\int\psi_k\,\dd\eta$ is constant in the read token. All other measures have infinite objective. The variational heads therefore agree with the original construction on every context, so the exact representation and its global stability are unchanged. This concerns the representing network constructed in that theorem, not a variational realization of every prescribed supported kernel.

\subsection{Projected Wasserstein regularization and SelfWarp reads}
\label{app:wasserstein-selfwarp}

\begin{proof}[Proof of Proposition~\ref{prop:wasserstein-calR-stable}]
    Write $F(\mu)=\calP(\supp\mu)$. Since $\sfE$ is compact, this is a nonempty compact subset of $\calP(\sfE)$. For each fixed $(\mu,z)$, the finite-valued cost is continuous and bounded in $w$, so its integral is continuous in $\kappa$. Continuous pushforward and continuity of $W_2$ on probabilities over compact $\sfX$ show that $J_{\mu,z}$ is continuous in $\kappa$. It therefore attains its minimum on $F(\mu)$.

    To obtain a Borel choice of minimizers, first observe that $F$ is lower hemicontinuous. Given $\kappa\in F(\mu)$ and an open neighborhood $U$ of $\kappa$, choose a finitely supported probability $\sum_i a_i\delta_{w_i}\in U$ with $w_i\in\supp\mu$. There are open neighborhoods $O_i$ of $w_i$ such that $\sum_i a_i\delta_{w_i'}\in U$ whenever $w_i'\in O_i$. Each $\mu(O_i)>0$. By the Portmanteau theorem, every sufficiently nearby $\mu'$ has $\mu'(O_i)>0$, and hence a support point $w_i'\in O_i$. Thus $F(\mu')\cap U\ne\emptyset$. In particular, $F$ is a measurable compact-valued correspondence.

    By the standing measurability convention, $c$ is jointly Borel. Parameter-dependent integration therefore makes $(\mu,z,\kappa)\mapsto J_{\mu,z}(\kappa)$ Borel; it is continuous in $\kappa$ for fixed $(\mu,z)$. The measurable minimum theorem gives a measurable nonempty compact-valued argmin correspondence. The measurable selection theorem \citep[Theorem~1]{himmelberg1969selection} supplies a Borel selector $K[\mu,z]$. Since $K[\mu,z]\in F(\mu)$, this is a supported Markov kernel satisfying \eqref{eq:objective-function-J}. No uniqueness or continuity of the selection is asserted.
\end{proof}

\begin{proposition}[Exact recovery of SelfWarp reads]
    \label{prop:wasserstein-selfwarp}
    Let $(\sfX,\rho_\sfX)$ be a compact metric space, let $\sfA\subset\R^m$ be compact, and set $\sfE=\sfX\times\sfA$ with spatial projection $\pi_\sfX$. Let $\lambda\in\calP(\sfX)$ have full support. Fix a Borel address map $\tau_h:\sfE\to\sfX$ and a constant $B> D:=\operatorname{diam}(\sfX)$. Use the regularizer \eqref{eq:wasserstein-calR} and the cost
    \begin{equation}
        \label{eq:wasserstein-selfwarp-cost}
        c_h(\mu,z,w)=B\rho_\sfX\bigl(\pi_\sfX(w),\tau_h(z)\bigr).
    \end{equation}
    For every continuous $f:\sfX\to\sfA$, write $j_f(y)=(y,f(y))$ and $\gamma_f=(j_f)_\#\lambda$. Then, for every $z\in\sfE$, the unique minimizer in \eqref{eq:objective-function-J} is
    \begin{equation}
        \label{eq:wasserstein-selfwarp-kernel}
        K_h[\gamma_f,z]=\delta_{j_f(\tau_h(z))}
        =\delta_{(\tau_h(z),f(\tau_h(z)))}.
    \end{equation}
    In particular, on a compact Euclidean domain or torus, take $\tau_h(x,a)=x+g_h(a)$, assuming the address lies in $\sfX$ or using periodic wrapping. With the value map $v_h(y,b)=V_hb$, the resulting read is
    \begin{equation}
        \label{eq:wasserstein-selfwarp-read}
        r_h(\gamma_f,(x,f(x)))=V_hf\bigl(x+g_h(f(x))\bigr),
    \end{equation}
    which is the continuum SelfWarp head of \citet[Appendix~C.2]{muser2026flowers}.
\end{proposition}

\begin{proof}
    Fix $f,z$, and write $a=\tau_h(z)$. Continuity of $f$ and full support of $\lambda$ give $\supp\gamma_f=j_f(\sfX)$, so $\kappa_a:=\delta_{j_f(a)}$ is feasible. For any feasible $\kappa$, set $\nu=(\pi_\sfX)_\#\kappa$, and let $q$ be an optimal coupling of $\nu$ and $\lambda$ for $W_2$. Since both distances below are bounded by $D$, the reverse triangle inequality gives
    $$
        \frac12\bigl(\rho_\sfX(y,t)^2-\rho_\sfX(a,t)^2\bigr)
        \ge -D\rho_\sfX(y,a).
    $$
    Integrating against $q(\dd y,\dd t)$ yields
    $$
        \frac12 W_2(\nu,\lambda)^2
        -\frac12 W_2(\delta_a,\lambda)^2
        \ge -D\int_\sfX\rho_\sfX(y,a)\,\nu(\dd y).
    $$
    Hence the objective corresponding to \eqref{eq:wasserstein-selfwarp-cost} satisfies
    \begin{equation}
        \label{eq:wasserstein-selfwarp-gap}
        J_{\gamma_f,z}(\kappa)-J_{\gamma_f,z}(\kappa_a)
        \ge (B-D)\int_\sfX\rho_\sfX(y,a)\,\nu(\dd y).
    \end{equation}
    As $B>D$, every minimizer has spatial marginal $\delta_a$. The graph $j_f(\sfX)$ contains exactly one token above $a$, so its only supported probability measure with that marginal is $\kappa_a$. This proves uniqueness and \eqref{eq:wasserstein-selfwarp-kernel}; integrating the value map proves \eqref{eq:wasserstein-selfwarp-read}. The cost is Borel and continuous in $w$, so every minimizing Markov kernel supplied by Proposition~\ref{prop:wasserstein-calR-stable} has the stated restriction to continuous graph contexts.
\end{proof}

\paragraph{Scope of the recovery.}
For a general context $\mu$, if $a=\tau_h(z)\in\pi_\sfX(\supp\mu)$, the same argument with $(\pi_\sfX)_\#\mu$ in place of $\lambda$ shows that the full set of minimizers is
$$
    \calP\bigl(\supp\mu\cap\pi_\sfX^{-1}(\{a\})\bigr).
$$
Thus the spatial cost alone does not select between multiple observed feature values at the same address. Proposition~\ref{prop:wasserstein-selfwarp} recovers the continuum graph-context reads; the finite-grid interpolation rows and boundary conventions are described separately in Section~\ref{sec:concrete-flowers}. No stability assertion for arbitrary noisy contexts follows from this exact-recovery statement.

\section{Approximation by uniformly discretization stable operators}
\label{sec:universal-approximation}

Uniform discretization stability does not prevent approximation of continuous operators on compact sets of input functions.

\begin{proposition}
    \label{prop:universal-approximation}
    Let $\calD \subset L^2(\sfX, \sfY)$ be compact. For every continuous $A: \calD \to L^2(\sfX, \sfY)$ and $\epsilon > 0$, there exists $\calN: \calP(\sfG) \times \sfX \to \sfY$, uniformly discretization stable on $\calP(\sfG)$, such that
    \begin{equation}
        \sup_{f \in \calD} \| \calN(\gamma_f, \cdot) - A[f] \|_{L^2(\sfX,\R^\dout)} < \epsilon.
    \end{equation}
\end{proposition}

\begin{proof}
    We approximate $A$ by weighted averages of finitely many of its outputs, with weights that depend Lipschitz continuously on the input graph measure. If $\calD$ is empty, any constant map suffices, so assume $\calD\ne\emptyset$.

    First, the graph embedding is continuous because coupling the graph measures at the same spatial point gives
    $$
        W_1(\gamma_f,\gamma_g)
        \le\int_\sfX |f(x)-g(x)|\,\lambda(\dd x)
        \le\|f-g\|_{L^2}.
    $$
    It is also injective. Since $\calD$ is compact, this embedding is therefore a homeomorphism onto the compact set
    $$
        \calK=\{\gamma_f:f\in\calD\}\subset\calP(\sfG).
    $$
    Consequently, the induced map $\widetilde A:\calK\to L^2(\sfX,\sfY)$, defined by $\widetilde A(\gamma_f)=A[f]$, is uniformly continuous. Choose $\delta>0$ such that
    $$
        W_1(\gamma_f,\gamma_g)<\delta
        \quad\Longrightarrow\quad
        \|A[f]-A[g]\|_{L^2}<\epsilon/2
        \qquad\text{for }f,g\in\calD.
    $$
    Choose a finite $\delta/2$-net $\mu_i=\gamma_{f_i}$, $i=1,\ldots,m$, in $\calK$. For each $i$, choose a Borel representative $h_i:\sfX\to\sfY$ of $A[f_i]$. Also fix $y_*\in\sfY$ and set $h_0(x)=y_*$.

    We now define the weights on all of $\calP(\sfG)$. Put
    $$
        b_i(\mu)=\max\{\delta-W_1(\mu,\mu_i),0\},
        \qquad
        B(\mu)=\max\left\{\sum_{i=1}^m b_i(\mu),\delta/2\right\},
    $$
    and set
    $$
        w_i(\mu)=\frac{b_i(\mu)}{B(\mu)}\quad(i=1,\ldots,m),
        \qquad
        w_0(\mu)=1-\sum_{i=1}^m w_i(\mu).
    $$
    All these weights are nonnegative and sum to one. They are globally Lipschitz: the functions $b_i$ are bounded and Lipschitz, and the Lipschitz denominator $B$ is bounded below by $\delta/2$. If $\mu\in\calK$, the net property gives $\sum_i b_i(\mu)\ge\delta/2$, so $w_0(\mu)=0$. Moreover, $w_i(\mu)>0$ implies $W_1(\mu,\mu_i)<\delta$.

    Define
    $$
        \calN(\mu,x)=\sum_{i=0}^m w_i(\mu)h_i(x).
    $$
    This map is Borel measurable and takes values in $\sfY$ by convexity. Since the $h_i$ are uniformly bounded and the weights are Lipschitz, there is a constant $C_\epsilon>0$ such that, for all $\mu,\nu\in\calP(\sfG)$,
    \begin{align*}
        \sup_{x\in\sfX}|\calN(\mu,x)-\calN(\nu,x)|
        &\le\sum_{i=0}^m\left(\sup_{x\in\sfX}|h_i(x)|\right)
            |w_i(\mu)-w_i(\nu)|\\
        &\le C_\epsilon\,W_1(\mu,\nu).
    \end{align*}
    Thus $\calN$ is uniformly discretization stable on $\calP(\sfG)$ with $\beta=1$; the constant $C_\epsilon$ may depend on the requested accuracy.

    Finally, fix $f\in\calD$. At $\mu=\gamma_f$, the weight $w_0$ vanishes and each active weight $w_i$ corresponds to an output $A[f_i]$ within $\epsilon/2$ of $A[f]$ in $L^2$. Therefore
    $$
        \|\calN(\gamma_f,\cdot)-A[f]\|_{L^2}
        \le\sum_{i=1}^m w_i(\gamma_f)\,
            \|A[f_i]-A[f]\|_{L^2}
        \le\epsilon/2.
    $$
    Taking the supremum over $f\in\calD$ gives the required bound, since $\epsilon/2<\epsilon$.
\end{proof}

The approximating map can also be chosen to be a stable Markov kernel network. For nonempty $\calD$, all measures in $\calK$ have spatial marginal $\lambda$, so $\calK$ is admissible by the same argument used in the operator-characterization corollary. Applying Theorem~\ref{thm:representation-2} to the restriction of the constructed map to $\calK\times\sfX$ gives a stable Markov kernel network with the same approximation guarantee. For empty $\calD$, a constant Markov kernel network suffices. This conclusion concerns the abstract architecture of Definition~\ref{def:MKN}.

\section{Concrete subclasses of Markov kernel networks}
\label{sec:concrete-subclasses}

We give explicit realizations of transformers, neural operators, DeepONets, and Flowers in Definition~\ref{def:MKN}. In each case, the Markov kernels specify where a head reads, the value maps specify what it reads, and the channel mixer combines the reads with the query token. The constructions reproduce the stated architectures exactly on their specified input contexts. They do not assert discretization stability: in particular, point evaluation may be unstable in the graph-Wasserstein metric.

\subsection{Conventions, context updates, and output queries}
\label{subsec:architecture-conventions}

We work with fixed parameters and finitely many layers on a compact metric spatial domain $\sfX$ with at least two points, as in the main text. Hidden tokens have the form $z=(x,a)\in\sfE_\ell=\sfX\times\sfA_\ell$, where $\sfA_\ell\subset\R^{d_\ell}$ is a compact box containing the feature values under consideration. Thus these are bounded-state realizations, rather than claims about an architecture on an entire unbounded function space. A pointwise input lifting is encoded as $q_\sfG(x,y)=(x,e^\theta(x,y))$. The coordinate $x$ is retained as a spatial label; it need not itself be supplied as a feature to every learned map. Continuum realizations using point reads below assume continuous hidden fields.

For a continuous hidden field $u:\sfX\to\sfA$ and a fixed probability measure $\lambda$ with $\supp\lambda=\sfX$, write
\begin{equation}
    \label{eq:architecture-graph-context}
    j_u(x)=(x,u(x)),\qquad \gamma_u=(j_u)_\#\lambda.
\end{equation}
If $\Gamma(\gamma_u,(x,u(x)))=(x,u^+(x))$, its context update is exactly $B_\Gamma(\gamma_u)=(\identity,u^+)_\#\lambda$. Likewise, for an empirical context $\mu=\sum_i\omega_i\delta_{(x_i,a_i)}$, with positive weights summing to one, the pushforward updates all features simultaneously while retaining the labels and weights:
$$
    B_\Gamma(\mu)=\sum_i\omega_i\delta_{\Gamma(\mu,(x_i,a_i))}.
$$
Empirical formulas below reproduce the corresponding discrete computation; they are not assertions that a quadrature sum equals its continuum integral.

\paragraph{Supported point reads.}
Continuity and full spatial support imply $\supp\gamma_u=j_u(\sfX)$. The graph law therefore determines $u$, and the addressed read
\begin{equation}
    \label{eq:architecture-graph-read}
    D_y[\gamma_u]:=\delta_{(y,u(y))},\qquad y\in\sfX,
\end{equation}
is supported on the context. For a finite context with distinct spatial labels, $D_{x_i}[\mu]=\delta_{(x_i,a_i)}$ is similarly valid at a supplied label. An arbitrary off-grid point requires a specified reconstruction rule. These reads are Borel maps on their stated domains. Indeed, on continuous graph contexts, $u\mapsto\gamma_u$ is a continuous injection from the Polish space $C(\sfX,\sfA)$ into $\calP(\sfE)$; its image and inverse are Borel, and point evaluation is jointly continuous in $(u,y)$. For empirical contexts, normalizing the restriction of $\mu$ to $\{y\}\times\sfA$ whenever this fiber has positive mass gives a jointly Borel supported read; it equals the stated Dirac measure when the feature at $y$ is unique.

A final pointwise projection $Q$ is consequently implemented by
\begin{equation}
    \label{eq:architecture-pointwise-decoder}
    \begin{aligned}
        q_\sfX(x)&=(x,a_*),&
        K_p[\mu,z]&=D_{\pi_\sfX z}[\mu],\\
        v_p(y,b)&=b,&
        m_p(z,r)&=Q(r),
    \end{aligned}
\end{equation}
where $a_*$ is any fixed feature vector in the final box. In particular, the encoder does not need access to the unknown output feature: the decoder kernel obtains it from the context. On a grid this realizes the projection at existing labels. Off-grid reconstruction requires its own supported reads and mixer, as illustrated for Flowers below.

\paragraph{Domains of the component maps.}
Each construction specifies the contexts on which equality with the original architecture is claimed. To obtain kernels on all of $\calP(\sfE)\times\sfE$, use $K[\mu,z]=\mu$ whenever a stated domain or mass condition fails, including an absent graph, sensor, or positive denominator. The relevant graph and fixed-grid context classes are Borel, so this gives supported Borel extensions. Values are bounded on the compact feature spaces. Since probability reads lie in the closed convex hulls of their value ranges, the mixers can be extended to arbitrary read vectors by clipping to bounding boxes and then clipping their outputs to sufficiently large feature boxes. Such extensions leave the stated realizations unchanged. For a prescribed compact convex output set $\sfY$, projection onto $\sfY$ likewise leaves represented outputs in $\sfY$ unchanged. None of these extensions is claimed to preserve stability automatically.

\subsection{Transformers}
\label{app:transformers}

We realize the coordinate-preserving attention and pointwise MLP layers of \citet[Section~4, Eqs.~(21)--(24)]{furuya2026functiongraph}. Here $\sfX\subset\R^{d_x}$, and hidden tokens have the form $z=(x,a)$ with $a\in\R^d$. The spatial coordinate is passed unchanged to the output, while the query, key, value, and pointwise feature maps may all depend on the full token. The input lifting can be the identity $q_\sfG(x,y)=(x,y)$, or a coordinate-preserving pointwise embedding $q_\sfG(x,y)=(x,e^\theta(x,y))$. No separate additive positional embedding is needed for this construction.

Suppress the layer index. Let $d$, $p$, and $d'$ be the feature dimensions before attention, after attention, and after the following MLP, respectively. For head $h$, let $k_h$ be the query/key dimension and $d_h$ the value dimension. Choose
\begin{equation}
    \begin{aligned}
        Q_h^\theta,K_{\mathrm{key},h}^\theta&\in\R^{k_h\times(d_x+d)},
        &V_h^\theta&\in\R^{d_h\times(d_x+d)},\\
        W_h^\theta&\in\R^{p\times d_h},
        &W_{\mathrm{loc}}^\theta&\in\R^{p\times(d_x+d)}.
    \end{aligned}
\end{equation}
The subscript $\mathrm{key}$ distinguishes the attention key matrix from a Markov kernel. The local matrix $W_{\mathrm{loc}}^\theta$ corresponds to $W^L$ in the cited construction. For $z=(x,a)$ and $w=(y,b)$, define
\begin{align}
    s_h^\theta(z,w)
        &:=\frac{\langle Q_h^\theta z,K_{\mathrm{key},h}^\theta w\rangle}{\sqrt{k_h}},
        \label{eq:appendix-attention-score}\\
    K_h[\mu,z](\dd w)
        &:=\frac{\exp s_h^\theta(z,w)}
        {\int_\sfE\exp s_h^\theta(z,\widetilde w)\,\mu(\dd\widetilde w)}\,
        \mu(\dd w),
        \qquad v_h(w):=V_h^\theta w.
        \label{eq:appendix-attention-kernel}
\end{align}
Each row is a probability measure supported on $\supp\mu$. The read $r_h$ has dimension $d_h$, and $\sum_hW_h^\theta r_h\in\R^p$ is the output projection of the concatenated head values. The matrices $W_h^\theta$ are learned channel maps; preservation of the spatial coordinate is imposed separately by the mixer. The scaling involves the query/key dimension $k_h$, which need not equal the value dimension $d_h$.

The coordinate-preserving pointwise MLP has the form
\begin{equation}
    F^\theta(x,t)=\bigl(x,H^\theta(x,t)\bigr),
    \qquad H^\theta:\R^{d_x+p}\to\R^{d'},
    \label{eq:appendix-transformer-ffn}
\end{equation}
where $H^\theta$ is a feed-forward MLP acting on the full position/feature pair. For example, one may take $H^\theta(x,t)=A_2^\theta\sigma(A_1^\theta(x,t)+b_1^\theta)+b_2^\theta$, with $A_1^\theta\in\R^{d_{\rm ff}\times(d_x+p)}$, $A_2^\theta\in\R^{d'\times d_{\rm ff}}$, $b_1^\theta\in\R^{d_{\rm ff}}$, $b_2^\theta\in\R^{d'}$, and coordinatewise ReLU activation $\sigma$. Combining attention with this pointwise map gives the channel mixer
\begin{equation}
    t=W_{\mathrm{loc}}^\theta z+\sum_{h=1}^HW_h^\theta r_h,
    \qquad
    m\bigl(z,r_1,\ldots,r_H\bigr)
        =\bigl(x,H^\theta(x,t)\bigr),\qquad z=(x,a).
    \label{eq:appendix-transformer-mixer}
\end{equation}
The attention map is $(x,a)\mapsto(x,t)$, and \eqref{eq:appendix-transformer-mixer} applies $F^\theta$ to its output. Since $F^\theta$ is independent of the context, the two operations form a single Markov kernel layer. If $p=d$, the choice $W_{\mathrm{loc}}^\theta(x,a)=a$ gives the usual feature residual connection. If also $d'=d$, an MLP residual can be written $H^\theta(x,t)=t+f^\theta(x,t)$. The displayed construction has no layer normalization or dropout, as in the cited attention/MLP formulas.

For $\mu=\sum_{j=1}^n\omega_j\delta_{z_j}$, with $z_j=(x_j,a_j)$, $\omega_j>0$, and $\sum_j\omega_j=1$, the heads give
\begin{equation}
    r_h(\mu,z_i)=\sum_{j=1}^n\alpha_{ij}^hV_h^\theta z_j,
    \qquad
    \alpha_{ij}^h=
    \frac{\omega_j\exp s_h^\theta(z_i,z_j)}
    {\sum_{q=1}^n\omega_q\exp s_h^\theta(z_i,z_q)}.
    \label{eq:appendix-empirical-attention}
\end{equation}
Uniform weights recover ordinary rowwise softmax; general weights give its quadrature-weighted counterpart. If $m(z_i,r_1,\ldots,r_H)=(x_i,a_i^+)$, the bulk update is
\begin{equation}
    B_\Gamma(\mu)=\sum_{i=1}^n\omega_i\delta_{(x_i,a_i^+)}.
\end{equation}
Thus every token is updated from the same previous context, and the pushforward retains its spatial coordinate and sampling weight. In particular, a graph context $\gamma_u$ is sent to $\gamma_{u^+}$, where $u^+(x)$ is the feature component of $\Gamma(\gamma_u,(x,u(x)))$. Reordering complete position/feature tokens merely reorders their outputs; the dependence of the learned maps on the spatial coordinates supplies the positional information.

On a finite sequence, causal and padding masks are included by multiplying the exponential in both the numerator and denominator of \eqref{eq:appendix-attention-kernel} by the prescribed mask $M_h(z,w)\in\{0,1\}$. Every represented query must have at least one permitted source token. This is again a supported probability row.

\paragraph{Readouts and propagated queries.}
For a final pointwise output map $O^\theta$, the supported graph read $D_x[\mu_L]$ from \eqref{eq:architecture-graph-read}, followed by the feature projection and $O^\theta$, realizes the continuum output $O^\theta(u_L(x))$; on an empirical context this is the native readout at supplied locations. A separate final attention decoder can use a context-independent query encoder $q_\sfX(x)=(x,\psi^\theta(x))\in\sfE_L$ and full-token scores and values
\begin{equation}
    s_{p,h}^\theta(z,w)
      =\frac{\langle Q_{p,h}^\theta z,K_{\mathrm{key},p,h}^\theta w\rangle}
      {\sqrt{k_{p,h}}},
    \qquad v_{p,h}(w)=V_{p,h}^\theta w,
    \qquad z=q_\sfX(x).
\end{equation}
Insert this score into \eqref{eq:appendix-attention-kernel} on $\sfE_L$. The decoder mixer combines these reads with the query token and applies the output projection. The query need not be an atom of the source context.

The fixed-query-value construction of \citet[Eq.~(25)]{furuya2026functiongraph} uses an additional query trajectory. With the identity input lifting, it starts from $z_0^{\mathrm q}=(x,a_*)$ and propagates
\begin{equation}
    z_{\ell+1}^{\mathrm q}=\Gamma_\ell(\mu_\ell,z_\ell^{\mathrm q}),
    \qquad
    \mu_{\ell+1}=\Gamma_\ell(\mu_\ell,\cdot)_\#\mu_\ell,
\end{equation}
with the query initialized independently of the source feature at $x$ and never inserted into the source context. Its final feature can differ from that of a source token at the same position. The identification of hidden layers above is exact; reproducing that additional query convention requires retaining its trajectory, rather than identifying it with either of the final-context readouts just described.

\subsection{Neural operators}
\label{sec:concrete-neural-operators}

On a common spatial domain, with the identity local coordinate map and a normalized integration measure, the nonlinear integral layers of \citet[Section~3]{kovachki2023neuraloperator} have the form
\begin{equation}
    \label{eq:no-layer}
    u^+(x)=\sigma\!\left(W^\theta u(x)+b^\theta(x)
    +\int_\sfX \kappa^\theta[u](x,y)u(y)\,\lambda(\dd y)\right).
\end{equation}
Here $u(x)\in\R^d$, the updated feature lies in $\R^p$, $W^\theta\in\R^{p\times d}$, and the integral kernel is matrix-valued. Its dependence on the current features, or on retained input channels, is permitted. In graph-context notation, write the integral as
\begin{equation}
    \label{eq:no-integral}
    I_i^\theta(\mu,z)
    =\sum_{j=1}^d\int_\sfE k_{ij}^\theta(\mu,z,w)b_j\,\mu(\dd w),
    \qquad z=(x,a),\quad w=(y,b).
\end{equation}
For example, $k^\theta(\gamma_u,(x,u(x)),(y,u(y)))=\kappa^\theta[u](x,y)$. We take $k^\theta:\calP(\sfE)\times\sfE\times\sfE\to\R^{p\times d}$ Borel measurable, allowing context dependence subject to the mass bound below. The signed matrix entries need not themselves be probability densities. A fixed finite integration measure $\nu=c\lambda$ is included by absorbing $c=\nu(\sfX)$ into the kernel.

Choose continuous $\phi_0,\phi_1:\sfX\to[0,1]$ and nonzero continuous $\rho_0,\rho_1:\sfX\to[0,\infty)$ such that
\begin{equation}
    \label{eq:no-completion-functions}
    \phi_0+\phi_1=1,\qquad \phi_s\rho_s=0,\qquad s\in\{0,1\}.
\end{equation}
Such functions exist because $\sfX$ has at least two points: choose a function with plateaus $0$ and $1$ on disjoint neighborhoods, its complement, and bumps inside their respective zero regions. The partition splits each value into two pieces. Each piece then has a region where additional probability mass contributes zero to its read.

\begin{lemma}[Markov realization of signed integral layers]
    \label{lem:no-markov-realization}
    Let the feature range be compact and fix the parameters $\theta$. Suppose that, on the represented contexts and queries, the positive and negative parts of every entry satisfy
    \begin{equation}
        \label{eq:no-mass-bound}
        a_{ij}^{\pm}(\mu,z)
        :=\int_\sfE (k_{ij}^\theta)^{\pm}(\mu,z,w)\,\mu(\dd w)
        \leq M
    \end{equation}
    for some fixed $M>0$. For the functions in \eqref{eq:no-completion-functions}, also assume $c_s(\mu):=\int_\sfE\rho_s(y)\,\mu(\dd w)>0$ for $s=0,1$. Then \eqref{eq:no-integral} is recovered exactly from $4pd$ scalar Markov read heads, each absolutely continuous with respect to $\mu$, by a linear channel mixer.
\end{lemma}

\begin{proof}
    Define the completion probabilities
    \begin{equation}
        \nu_s[\mu](\dd w):=\frac{\rho_s(y)}{c_s(\mu)}\,\mu(\dd w).
    \end{equation}
    For the full-support graph contexts of Section~\ref{subsec:architecture-conventions}, $c_s(\gamma_u)=\int_\sfX\rho_s\,\dd\lambda>0$ is independent of $u$.

    Index the heads by $h=(i,j,s,\epsilon)$, where $\epsilon\in\{+,-\}$. Set
    \begin{align}
        \label{eq:no-markov-kernel}
        K_{ij,s,\epsilon}[\mu,z](\dd w)
        &:=\frac{(k_{ij}^\theta)^\epsilon(\mu,z,w)}{M}\,\mu(\dd w)
        +\left(1-\frac{a_{ij}^\epsilon(\mu,z)}{M}\right)\nu_s[\mu](\dd w),\\
        \label{eq:no-value-map}
        v_{ij,s,\epsilon}(w)&:=\phi_s(y)b_j.
    \end{align}
    The two nonnegative summands in \eqref{eq:no-markov-kernel} have masses $a_{ij}^\epsilon/M$ and $1-a_{ij}^\epsilon/M$, respectively. Hence $K_{ij,s,\epsilon}$ is a probability measure supported on $\supp\mu$. The bounded value map depends only on $w$. Since it vanishes $\nu_s[\mu]$-almost everywhere, its read is
    \begin{equation}
        r_{ij,s,\epsilon}(\mu,z)
        =\frac1M\int_\sfE (k_{ij}^\theta)^\epsilon(\mu,z,w)
        \phi_s(y)b_j\,\mu(\dd w).
    \end{equation}
    Summing over the partition and subtracting the signed parts gives
    \begin{equation}
        \label{eq:no-reconstruction}
        I_i^\theta(\mu,z)
        =M\sum_{j=1}^d\sum_{s=0}^1
        \bigl(r_{ij,s,+}(\mu,z)-r_{ij,s,-}(\mu,z)\bigr).
    \end{equation}
\end{proof}

Thus the channel mixer for \eqref{eq:no-layer} is
\begin{equation}
    \label{eq:no-mixer}
    \begin{aligned}
        m^\theta\bigl((x,a),r\bigr)
        &:=\left(x,\sigma\!\left(W^\theta a+b^\theta(x)+J(r)\right)\right),\\
        J_i(r)&:=M\sum_{j=1}^d\sum_{s=0}^1(r_{ij,s,+}-r_{ij,s,-}).
    \end{aligned}
\end{equation}
In particular, the masses $a_{ij}^{\pm}$ remain inside the kernels; the mixer does not require them as additional inputs. Preserving $x$ makes the context pushforward precisely the graph law of $u^+$. Applying this construction at every layer, together with the lifting of Section~\ref{subsec:architecture-conventions} and the decoder \eqref{eq:architecture-pointwise-decoder}, recovers the complete network whenever the represented hidden fields remain continuous. Usual continuous local kernels and pointwise maps have this property. Borel kernels alone need not preserve it. Any original-input channels used by the kernels are retained through the layers.

The pointwise bound $|k_{ij}^\theta|\leq M$ is sufficient for \eqref{eq:no-mass-bound}, but a uniform row-mass bound is all the proof uses. For the usual continuous local kernel parameterizations, compact feature ranges give a pointwise bound. General dependence on the entire context requires the mass bound separately. The same identities hold for a weighted empirical context $\mu=\sum_n\omega_n\delta_{(x_n,u_n)}$ whenever \eqref{eq:no-mass-bound} holds and $\sum_n\omega_n\rho_s(x_n)>0$ for both $s$. They then reproduce the corresponding quadrature layer exactly. The completion condition holds eventually along spatial quadratures converging weakly to $\lambda$, but need not hold on arbitrary coarse samples. These are representation statements; no discretization-stability assertion follows from the signed decomposition alone.

\subsubsection{Fourier neural operators}
\label{sec:concrete-fno}

The finite-mode spectral layers of Fourier neural operators \citep{li2021fourier} admit a simpler realization. Let $\sfX=\mathbb T^{d_x}$ with normalized volume $\lambda$, and let $\Lambda\subset\mathbb Z^{d_x}$ be the finite set of retained modes. With $R_k^\theta\in\mathbb C^{p\times d}$, their spectral contribution is
\begin{equation}
    \label{eq:fno-spectral-layer}
    S^\theta[u](x)=\sum_{k\in\Lambda}e^{2\pi\mathrm i k\cdot x}
    R_k^\theta\widehat u(k),
    \qquad
    \widehat u(k)=\int_\sfX e^{-2\pi\mathrm i k\cdot y}u(y)\,\lambda(\dd y).
\end{equation}
Take the usual conjugate symmetry of the modes and multipliers so that $S^\theta[u]$ is real. A single Markov head with
\begin{equation}
    \label{eq:fno-markov-head}
    K[\mu,z]=\mu,
    \qquad
    v(y,b)=\bigl(\cos(2\pi k\cdot y)b,\,
    \sin(2\pi k\cdot y)b\bigr)_{k\in\Lambda}
\end{equation}
returns $2d|\Lambda|$ real Fourier moments $r=(C_k,S_k)_{k\in\Lambda}$, with $\widehat u(k)=C_k-\mathrm i S_k$. The channel mixer is
\begin{equation}
    \label{eq:fno-mixer}
    \begin{aligned}
        m^\theta((x,a),r)
        &:=\left(x,\sigma\!\left(W^\theta a+b^\theta(x)+F^\theta(x,r)\right)\right),\\
        F^\theta(x,r)
        &:=\operatorname{Re}\sum_{k\in\Lambda}e^{2\pi\mathrm i k\cdot x}
        R_k^\theta(C_k-\mathrm i S_k).
    \end{aligned}
\end{equation}
Complex notation in this expression abbreviates a real linear map on the sine and cosine moments. Thus the Markov kernel performs an unweighted context read, while the value map collects the Fourier moments and the mixer applies the learned multipliers and synthesizes the result at $x$. The number of read channels depends on the retained modes, not on the spatial resolution.

On a uniform periodic grid, $\mu=N^{-1}\sum_{n=1}^N\delta_{(x_n,u_n)}$ gives normalized discrete Fourier coefficients, and \eqref{eq:fno-mixer} agrees with the corresponding FFT layer when the forward and inverse normalization conventions are matched. Retained modes must be represented with the same grid-frequency and conjugate-symmetry conventions. Other positive quadrature weights give the associated quadrature Fourier layer; equality to the continuum coefficients requires a separate quadrature argument. The formulas use the actual normalized-volume integration measure, or its declared discrete quadrature. Other fixed domain volumes or transform conventions require their known normalization factors. The local term, pointwise maps, and final projection are handled as in Section~\ref{subsec:architecture-conventions}.

\subsection{DeepONets}
\label{app:deeponets}

A fixed-sensor DeepONet pairs coefficients from a branch network with features from a trunk network \citep[Section 2.1]{lu2021deeponet}. For scalar output, sensors $s_1,\ldots,s_S\in\sfX$, and branch/trunk width $P$, its map is
\begin{equation}
    \calN(\gamma_f,x)
      =\left\langle
           B^\theta\bigl(f(s_1),\ldots,f(s_S)\bigr),T^\theta(x)
        \right\rangle+b_0^\theta,
    \label{eq:appendix-deeponet}
\end{equation}
where $B^\theta:\sfA^S\to\R^P$ and $T^\theta:\sfX\to\R^P$ are the branch and trunk maps. Vector outputs are obtained by applying this construction to each output component. We use the common input/output domain of Definition~\ref{def:MKN}.

Take the identity lifting $q_\sfG(x,a)=(x,a)$. To satisfy $L>0$ in that definition, use one identity context layer: $K_0[\mu,z]=\mu$, $v_0\equiv0$, and $m_0(z,r)=z$ give $B_0(\mu)=\mu$. All sensor extraction and branch--trunk synthesis can then occur in the decoder. Using the supported evaluation rows \eqref{eq:architecture-graph-read}, choose
\begin{equation}
    K_{p,j}[\mu,(x,a)]=D_{s_j}[\mu],
    \qquad v_{p,j}(y,b)=b,
    \qquad j=1,\ldots,S.
\end{equation}
On a continuous graph context $\gamma_f$, the $j$th read is exactly $r_j=f(s_j)$. The decoder mixer
\begin{equation}
    m_p\bigl((x,a),r_1,\ldots,r_S\bigr)
       =\langle B^\theta(r_1,\ldots,r_S),T^\theta(x)\rangle+b_0^\theta
\end{equation}
therefore gives \eqref{eq:appendix-deeponet}. It ignores the placeholder feature $a$ and uses $x$ in the trunk. Signed branch and trunk outputs are permitted because the probability constraint applies to the read measures.

For an empirical input, the same realization is exact when every sensor location occurs in the context with its observed value. The output query $x$ may be any point of $\sfX$, since its dependence enters through the trunk and does not require a sample at $x$. If a sensor is absent, its value requires a declared reconstruction rule. The fixed-sensor realization itself provides no such reconstruction and no uniform graph-$W_1$ stability assertion.

\subsection{Flowers}
\label{sec:concrete-flowers}

The SelfWarp heads, residual blocks, and multiscale composition of Flowers are specified in \citet[Appendix C]{muser2026flowers}. We first represent a block using the graph-context conventions of Section~\ref{subsec:architecture-conventions}, with $u:\sfX\to\R^{C_{\mathrm{in}}}$ continuous and bounded. Here $\sfX$ is a compact Euclidean domain or a torus. Let $C_{\mathrm{out}}=HC_h$, and write the value projection in head blocks as $V_h\in\R^{C_h\times C_{\mathrm{in}}}$. The displacement maps $g_h:\R^{C_{\mathrm{in}}}\to\R^{d_x}$ use all input channels. For $z=(x,a)$, let $\tau_h(z)=x+g_h(a)$, interpreted with periodic wrapping, or assume that this address lies in $\sfX$. Set
\begin{equation}
    \label{eq:flowers-supported-warp}
    K_h[\gamma_u,z]=D_{\tau_h(z)}[\gamma_u]=\delta_{(\tau_h(z),u(\tau_h(z)))},
    \qquad v_h(y,b)=V_hb.
\end{equation}
Then $r_h(\gamma_u,(x,u(x)))=V_hu(x+g_h(u(x)))$. Thus the mixer
$$
    m_{\mathrm{warp}}((x,a),r_1,\ldots,r_H)
    =\left(x,\,Pa+\bigoplus_{h=1}^H r_h\right)
$$
realizes SelfWarp plus its pointwise residual projection $P\in\R^{C_{\mathrm{out}}\times C_{\mathrm{in}}}$. For affine projections, value biases are included in $v_h$ and residual biases in the mixer. The output retains the target coordinate $x$. Since $\lambda$ has full support, the point selected in \eqref{eq:flowers-supported-warp} belongs to $\supp\gamma_u$.

\paragraph{Normalization.}
Write $b$ for the field produced by this first layer. GroupNorm uses moments of $b$ across space and channel groups, so it is a second Markov kernel layer. For every group $G\subset\{1,\ldots,C_{\mathrm{out}}\}$, take $K_G[\gamma_b,z]=\gamma_b$ and
$$
    v_G(y,b)=\left(\frac1{|G|}\sum_{c\in G}b_c,
                      \frac1{|G|}\sum_{c\in G}b_c^2\right).
$$
The resulting read gives the group mean $M_G$ and second moment $S_G$. Its mixer preserves $x$ and maps each feature channel $c\in G$ to
\begin{equation}
    \label{eq:flowers-groupnorm}
    \operatorname{GELU}\!\left(
        \gamma_c\frac{b_c-M_G}{\sqrt{\max\{S_G-M_G^2,0\}+\varepsilon}}+\beta_c
    \right),\qquad \varepsilon>0.
\end{equation}
For actual moment reads $S_G-M_G^2\geq0$; the maximum only extends the mixer to arbitrary read vectors. This reproduces the order residual sum, GroupNorm, then GELU. On a grid use $K_G=n^{-1}\sum_{j=1}^n\delta_{z_j}$ to obtain the required uniform spatial average regardless of context weights. When supplied as input, conditioning variables for FiLM can be retained in the token and used to determine $\gamma_c,\beta_c$. Stacking these two-layer realizations and using the decoder \eqref{eq:architecture-pointwise-decoder} gives the corresponding single-scale network, including its pointwise lifting and final projection.

\paragraph{Interpolation and boundaries.}
On a fixed grid with tokens $z_j=(x_j,a_j)$, let $b_j(y)\geq0$ be the sampler's multilinear interpolation weights. For constant-preserving boundary conventions, the implemented warp has
$$
    K_h[\mu,z]=\sum_j b_j(\tau_h(z))\delta_{z_j},
    \qquad \sum_j b_j(\tau_h(z))=1.
$$
These rows are supported on the sampled tokens. With zero padding, the sum $s_h(z):=\sum_j b_j(\tau_h(z))$ can be less than one. For $s_h(z)>0$, divide this row by $s_h(z)$ and multiply the resulting read by $s_h(z)$ in $m_{\mathrm{warp}}$; when $s_h(z)=0$, choose $K_h[\mu,z]=\mu$ and multiply the read by zero. The factor $s_h(z)$ is determined by the fixed sampler geometry and the local source address, so the mixer can compute it from $z$. This implements zero padding without adding unsupported zero-valued tokens.

\paragraph{The full multiscale network.}
The U-Net in \citet[Appendix C.4]{muser2026flowers} additionally uses strided and transposed convolutions and stored encoder skips. To represent it on its finite grid hierarchy $\Lambda_0,\ldots,\Lambda_J$, retain every finest-grid label $i\in\Lambda_0$ throughout the MKN. Choose surjections $c_\ell:\Lambda_0\to\Lambda_\ell$, with $c_0=\identity$, and representatives $\iota_\ell(y)\in c_\ell^{-1}(y)$. At scale $\ell$, token $i$ stores the active feature at $c_\ell(i)$ and the previously saved skip features at their corresponding coordinates. Warp addresses use the physical coordinate $c_\ell(i)$ at that scale; the original label $i$ is retained only to identify the carrier token. A read of a coarse site $y$ selects the existing token $z_{\iota_\ell(y)}$ and extracts its active channels. Consequently, each finite convolution stencil is represented by Dirac heads followed by the original, possibly signed, filter matrices in the mixer; the same construction realizes transposed convolutions and changes the active scale while retaining all finest-grid labels. Out-of-domain stencil terms receive zero coefficients. Coarse-grid interpolation uses the preceding formula with these representative tokens. GroupNorm uses the probability row
$$
    K_{\mathrm{norm},\ell}[\mu,z]
    =\frac1{|\Lambda_\ell|}\sum_{y\in\Lambda_\ell}
      \delta_{z_{\iota_\ell(y)}},
$$
which gives exactly the uniform coarse-grid moments, irrespective of the sizes of the fibers of $c_\ell$. Convolutional biases and the ReLU following each down/up convolution are applied in the mixer. Saved skip slots are copied unchanged through the active-channel operations; copying and concatenation are pointwise mixers, performed at the original architecture's stages. The encoder is $q_\sfG(x,a)=(x,\operatorname{Lift}(a\oplus x))$, with empty skip slots appended; the final supported point read at a finest-grid query, followed by the original pointwise projection, is $\Gamma_p$. Hence the complete finite-grid Flowers architecture is an MKN. Retaining the finest-grid labels avoids requiring a pushforward layer to create new tokens during upsampling. These exact realizations do not by themselves assert discretization stability of the learned point reads or of the multiscale network.

\section{Hybrid architectures through variational Markov reads}
\label{app:s4-proofs}

We construct a convex family of regularizers that recovers attention, uniform reads, and continuum SelfWarp at different parameter values. The read kernels satisfy a common H\"older bound on a fixed class of regular graph contexts. We also give a grid approximation bound and explain why the graph-context restriction is needed.

\subsection{Graph contexts and costs}
\label{app:s4-design-space}
Let $\sfX=[0,1]^d$, with $d\ge1$ and normalized Lebesgue probability measure $\lambda=\dd x$. Let $\sfV$ be a nonempty compact convex subset of a Euclidean space, and put $\sfE=\sfX\times\sfV$ with its Euclidean product metric. For $L_u\ge0$, define
$$
    \mathcal U_{L_u}:=\{u:\sfX\to\sfV:\operatorname{Lip}(u)\le L_u\},
    \qquad j_u(y):=(y,u(y)),\qquad \gamma_u:=(j_u)_\#\lambda.
$$
Every probability measure supported on $\supp\gamma_u=j_u(\sfX)$ has the unique form $\kappa=(j_u)_\#\nu$, with $\nu\in\calP(\sfX)$. Consequently,
$$
    \tfrac12W_2^2\bigl((\pi_\sfX)_\#\kappa,(\pi_\sfX)_\#\gamma_u\bigr)
    =\tfrac12W_2^2(\nu,\lambda),
    \qquad \operatorname{KL}(\kappa\Vert\gamma_u)=\operatorname{KL}(\nu\Vert\lambda).
$$
The KL identity follows because $j_u$ is a homeomorphism onto its graph; it also holds when the value is infinite.

To specify a common cost class, set
$$
    \mathcal T_{L_u}:=\{(\gamma_u,z,j_u(y)):
                         u\in\mathcal U_{L_u},\ z\in\sfE,\ y\in\sfX\}.
$$
Fix $M,L_c,L_y\ge0$ and $\beta\in(0,1]$. Let $\mathfrak C$ consist of the continuous real costs on $\mathcal T_{L_u}$ satisfying
\begin{equation}
    \label{eq:s4-costclass}
    \begin{aligned}
        |c(\gamma_u,z,j_u(y))|&\le M,\\
        |c(\gamma_u,z,j_u(y))-c(\gamma_v,z',j_v(y'))|
        &\le L_c\bigl(W_1(\gamma_u,\gamma_v)+\rho_\sfE(z,z')\bigr)^\beta
             +L_y|y-y'|.
    \end{aligned}
\end{equation}
\subsection{A stable Wasserstein--KL family}
Fix $\tau>0$. For $\alpha\in[0,1]$, define
\begin{equation}
    \label{eq:s4-spatial}
    J_{\alpha,c;u,z}(\nu)
    :=\int_\sfX c(\gamma_u,z,j_u(y))\,\nu(\dd y)
      +\frac{1-\alpha}{2}W_2^2(\lambda,\nu)
      +\alpha\tau\operatorname{KL}(\nu\Vert\lambda),
    \qquad \nu\in\calP(\sfX).
\end{equation}
At $\alpha=0$ the entropy term is omitted, rather than interpreted as $0\cdot(+\infty)$.

\begin{theorem}[A convex class of stable variational reads]
    \label{thm:s4-main}
    For every $c\in\mathfrak C$, $\alpha\in[0,1]$, $u\in\mathcal U_{L_u}$, and $z\in\sfE$, the objective \eqref{eq:s4-spatial} has a unique minimizer $\nu_{\alpha,c;u,z}$. Define the supported graph-token read
    \begin{equation}
        \label{eq:s4-routing}
        K_{\alpha,c}[\gamma_u,z]:=(j_u)_\#\nu_{\alpha,c;u,z}.
    \end{equation}
    Set
    $$
        \eta:=\min\left\{\frac\beta3,\frac1{d+1}\right\},
        \qquad \omega(0):=0,\qquad
        \omega(h):=h\log(e/h)\quad(0<h\le1).
    $$
    There is a constant $A$, depending only on $d,L_u,M,L_c,L_y,\beta,\tau$, and $\operatorname{diam}(\sfV)$, such that
    \begin{equation}
        \label{eq:s4-graph-stability}
        \begin{aligned}
            &W_1\bigl(K_{\alpha,c}[\gamma_u,z],
                      K_{\widetilde\alpha,\widetilde c}[\gamma_v,z']\bigr)\\
            &\quad\le A\left[
                \bigl(W_1(\gamma_u,\gamma_v)+\rho_\sfE(z,z')\bigr)^\eta
                +\left(\|c-\widetilde c\|_{\infty,\mathcal T_{L_u}}
                     +\omega(|\alpha-\widetilde\alpha|)\right)^{1/3}
            \right].
        \end{aligned}
    \end{equation}
    In particular, each fixed design defines a H\"older-stable Borel Markov kernel on this graph-context class. The design space $\mathfrak C\times[0,1]$ is convex and compact for uniform convergence of costs and ordinary convergence of $\alpha$.
\end{theorem}

The exponents are not claimed optimal. The estimate controls read kernels on the stated graph class; applying it through multiple layers requires preservation of the spatial marginal and the uniform graph regularity. It does not supply the all-context kernel bound in Definition~\ref{def:stable-markov-kernel-network}.

\subsection{Proof of Theorem~\ref{thm:s4-main}}
The main step is cubic transport growth under \emph{linear interpolation of measures}; entropy adds quadratic growth. Together they control the spatial minimizer uniformly. We then lift to graph tokens and compare coefficients using finite-entropy histograms, since a point read itself has infinite entropy.

Throughout this appendix, $D=\sqrt d$ and spatial Wasserstein distances use Euclidean distance on $\sfX=[0,1]^d$. For the two penalties, write $F(\nu)=\tfrac12W_2^2(\lambda,\nu)$ and $H(\nu)=\operatorname{KL}(\nu\Vert\lambda)$. Set
\begin{equation}\label{eq:s4-app-constants}
 q_d=(12dD)^{-1},\qquad
 m_{d,\tau}=\min\{q_d,\,2\tau/D^3\}>0.
\end{equation}
For the spatial proof, $c\in C(\sfX)$ denotes a single continuous function of position, and $\nu_{\alpha,c}$ denotes its minimizing spatial measure. The corresponding token read is its pushforward through $j_u$. Write
$$
 J_{\alpha,c}(\nu)=\int_\sfX  c\,\dd\nu+(1-\alpha)F(\nu)+\alpha\tau H(\nu),
 \qquad V(c)=\min_\nu J_{0,c}(\nu).
$$
We use standard compact Kantorovich duality and standard Wasserstein facts; the estimates needed below are proved explicitly.

\subsubsection*{Transport provides quantitative growth}
\paragraph{Growth at zero entropy.}
For every $c\in C(\sfX)$, $J_{0,c}$ has a unique minimizer $\nu_c$. For every $\nu\in\mathcal P(\sfX)$,
\begin{equation}\label{eq:s4-transport-growth}
 J_{0,c}(\nu)-V(c)\ge q_dW_1(\nu,\nu_c)^3.
\end{equation}
Consequently, if $\nu_b=(1-b)\nu_0+b\nu_1$ and $r=W_1(\nu_0,\nu_1)$, then
\begin{equation}\label{eq:s4-transport-convexity}
 (1-b)F(\nu_0)+bF(\nu_1)-F(\nu_b)
 \ge q_db(1-b)\bigl(b^2+(1-b)^2\bigr)r^3.
\end{equation}
\begin{proof}
\emph{Identify the optimizer.} Define on all of $\mathbb R^d$
$$
 \Phi_c(x)=\sup_{y\in \sfX}\{x\cdot y-\tfrac12|y|^2-c(y)\}.
$$
This convex Lipschitz function is differentiable almost everywhere, and its gradients lie in $\sfX$. At a differentiability point its maximizing destination is unique and equals $T_c(x)=\nabla\Phi_c(x)$. Define $T_c$ arbitrarily on the null exceptional set. For every source--destination pair,
$$
 \tfrac12|x-y|^2+c(y)\ge\tfrac12|x|^2-\Phi_c(x).
$$
The coupling $(\mathrm{id},T_c)_\#\lambda$ attains equality. Integrating against an optimal coupling for any candidate row gives
\begin{equation}\label{eq:s4-free-map}
 \nu_c=(T_c)_\#\lambda,\qquad
 V(c)=\int_\sfX \bigl(\tfrac12|x|^2-\Phi_c(x)\bigr)\,\dd\lambda.
\end{equation}
Every minimizing coupling must have destination $T_c(x)$ for almost every source, proving uniqueness.

\emph{Control the response to a cost perturbation.} If $\Phi,\Psi$ are smooth convex functions with gradients in $\sfX$, put $u=\Phi-\Psi$ and $e=\|u\|_{L^\infty(\sfX)}$. Integration by parts on the cube gives
$$
 \int_\sfX |\nabla u|^2\,\dd x
 =\int_{\partial \sfX}u\,\partial_nu\,\dd S
   -\int_\sfX u(\Delta\Phi-\Delta\Psi)\,\dd x\le4de.
$$
Indeed, each gradient component belongs to $[0,1]$, so the boundary integral of $|\partial_nu|$ is at most $2d$. Convexity and the divergence theorem give $0\le\int_\sfX \Delta\Phi\le d$ and the same bound for $\Psi$. For $\Phi_c,\Phi_{\widetilde c}$, global mollification preserves convexity, the gradient range, and the bound $\|\Phi_c-\Phi_{\widetilde c}\|_{L^\infty(\mathbb R^d)}\le\|c-\widetilde c\|_\infty$. Dominated convergence of the mollified gradients therefore yields
\begin{equation}\label{eq:s4-zero-response}
 W_1(\nu_c,\nu_{\widetilde c})\le W_2(\nu_c,\nu_{\widetilde c})
 \le 2\sqrt d\,\|c-\widetilde c\|_\infty^{1/2}.
\end{equation}

\emph{Turn response control into growth.} Set $r=W_1(\nu,\nu_c)$ and choose a $1$-Lipschitz Kantorovich--Rubinstein test $h$ with $\int h\,\dd(\nu-\nu_c)=r$ and $\|h\|_\infty\le D$. Testing the minimizers at neighboring costs in each other's objectives, then using \eqref{eq:s4-zero-response}, gives
$$
 \frac{\dd}{\dd t}V(c-th)=-\int h\,\dd\nu_{c-th}.
$$
The derivative is continuous by \eqref{eq:s4-zero-response}. Integrating it and using $\operatorname{Lip}(h)\le1$ gives
$$
 V(c-th)\ge V(c)-t\int h\,\dd\nu_c
                  -\tfrac43\sqrt{dD}\,t^{3/2}.
$$
On the other hand, $V(c-th)\le J_{0,c}(\nu)-t\int h\,\dd\nu$. Hence
$$
 J_{0,c}(\nu)-V(c)\ge tr-\tfrac43\sqrt{dD}\,t^{3/2}.
$$
Choosing $t=r^2/(4dD)$ proves \eqref{eq:s4-transport-growth}; the case $r=0$ follows from minimality.

Finally, compact Kantorovich duality supplies a continuous destination potential $b_*$ supporting $F$ at $\nu_b$: $F(\nu)\ge F(\nu_b)+\int b_*\,\dd(\nu-\nu_b)$. Thus $\nu_b$ minimizes $J_{0,-b_*}$. Apply \eqref{eq:s4-transport-growth} to $\nu_0,\nu_1$ and add with weights $1-b,b$. The linear terms cancel, while the dual norm formula for $W_1$ gives
$W_1(\nu_0,\nu_b)=br$ and $W_1(\nu_1,\nu_b)=(1-b)r$. This proves \eqref{eq:s4-transport-convexity}.
\end{proof}

\subsubsection*{Uniform control of the spatial minimizers}
Existence for every $\alpha$ follows from weak lower semicontinuity on compact $\mathcal P(\sfX)$ and the finite competitor $\lambda$. When the endpoint entropies are finite, linear interpolation has the identity
$$
 (1-b)H(\nu_0)+bH(\nu_1)-H(\nu_b)
 =(1-b)\operatorname{KL}(\nu_0\Vert\nu_b)
       +b\operatorname{KL}(\nu_1\Vert\nu_b).
$$
Pinsker's inequality and $W_1\le(D/2)\|\cdot\|_{\mathrm{TV}}$, with $\|\cdot\|_{\mathrm{TV}}$ the total variation norm, bound this below by $2b(1-b)W_1(\nu_0,\nu_1)^2/D^2$. Combine this with \eqref{eq:s4-transport-convexity}, take $\nu_0$ to be a minimizer, divide the interpolation deficit by $b$, and let $b\downarrow0$. For every finite-objective competitor, with $r=W_1(\nu,\nu_{\alpha,c})$, this proves
\begin{equation}\label{eq:s4-uniform-growth}
 J_{\alpha,c}(\nu)-J_{\alpha,c}(\nu_{\alpha,c})
 \ge(1-\alpha)q_dr^3+\frac{2\alpha\tau}{D^2}r^2
 \ge m_{d,\tau}r^3.
\end{equation}
At $\alpha=0$ no entropy assertion is used. The last inequality uses $r\le D$; it also proves uniqueness. Apply \eqref{eq:s4-uniform-growth} in both directions to the minimizers for $c$ and $\widetilde c$. Adding cancels the regularizer, so that
\begin{equation}\label{eq:s4-cost-response}
 W_1(\nu_{\alpha,c},\nu_{\alpha,\widetilde c})
 \le m_{d,\tau}^{-1/3}\|c-\widetilde c\|_\infty^{1/3},
 \qquad 0\le\alpha\le1.
\end{equation}
This proves the spatial estimate with no positive lower bound on the entropy coefficient.

\paragraph{Continuity in the entropy coefficient.}
If $\operatorname{Lip}(c)\le L_y$, then
\begin{equation}\label{eq:s4-coefficient}
 W_1(\nu_{\alpha,c},\nu_{\widetilde\alpha,c})
 \le C_{d,L_y,\tau}\,\omega(|\alpha-\widetilde\alpha|)^{1/3}
 \quad\text{for all }\alpha,\widetilde\alpha\in[0,1].
\end{equation}
\begin{proof}
\emph{Approximate the singular endpoint.} Divide $\sfX$ into $q^d$ cubes and let $S_q\nu$ spread each cell's mass uniformly within that cell. Then
\begin{equation}\label{eq:s4-histogram}
 W_1(S_q\nu,\nu)\le D/q,\qquad
 H(S_q\nu)\le d\log q,\qquad H(S_q\nu)\le H(\nu)
\end{equation}
when the last right-hand side is finite. The entropy statements follow from the cell probabilities and Jensen's inequality. Also
\begin{equation}\label{eq:s4-F-Lip}
 |F(\nu)-F(\nu')|\le D W_1(\nu,\nu').
\end{equation}
To see this, glue an optimal coupling from $\lambda$ to $\nu$ to a coupling between $\nu$ and $\nu'$, and use that $|x-y|^2/2$ is $D$-Lipschitz in $y$ on $\sfX$; then interchange the targets.

Set $C_0=D(L_y+D)$ and $V(c)=J_{0,c}(\nu_c)$. Testing $S_q\nu_c$ in the $\alpha$-problem, using $0\le F\le D^2/2$ and $H\ge0$, gives
\begin{equation}\label{eq:s4-boundary-comparison}
 \begin{aligned}
 J_{0,c}(\nu_{\alpha,c})-V(c)+\alpha\tau H(\nu_{\alpha,c})
 \le C_0/q+\alpha\tau d\log q+\alpha D^2/2.
 \end{aligned}
\end{equation}
For $\alpha>0$, take $q=\lceil1/\alpha\rceil$. The transport growth estimate and the nonnegativity of $J_{0,c}-V(c)$ imply, respectively,
\begin{align}
 W_1(\nu_{\alpha,c},\nu_c)^3
 &\le \frac{\alpha}{q_d}
       \bigl(C_0+D^2/2+\tau d\log(2/\alpha)\bigr),\label{eq:s4-boundary-rate}\\
 H(\nu_{\alpha,c})
 &\le B_0+d\log(2/\alpha),\qquad
 B_0=(C_0+D^2/2)/\tau.\label{eq:s4-entropy-bound}
\end{align}

\emph{Compare two positive coefficients.} By symmetry assume $0\le\alpha\le\widetilde\alpha\le1$, and put $h=\widetilde\alpha-\alpha$. For $\alpha\ge h>0$, write $J_{\widetilde\alpha,c}=J_{\alpha,c}+h(\tau H-F)$ and test each minimizer in the other's objective. Equation~\eqref{eq:s4-uniform-growth} bounds the sum of the two gaps below by $2m_{d,\tau}W_1(\nu_{\alpha,c},\nu_{\widetilde\alpha,c})^3$. The sum equals
$$
 h\left[(\tau H-F)(\nu_{\alpha,c})-(\tau H-F)(\nu_{\widetilde\alpha,c})\right]
 \le h\left[\tau B_0+\tau d\log(2/h)+D^2/2\right],
$$
by \eqref{eq:s4-entropy-bound}. This proves the required modulus in this regime. If $0\le\alpha<h$ and $h<1/4$, both coefficients are below $2h$; compare each row to $\nu_c$ using \eqref{eq:s4-boundary-rate}. If $h\ge1/4$, use $W_1\le D$. The case $h=0$ is immediate.
\end{proof}

\subsubsection*{From positions to tokens and completion of the proof}
First control the graph lift. Put $\delta=W_1(\gamma_u,\gamma_v)$ and $L_j=\sqrt{1+L_u^2}$. A coupling of the graph measures projects to a coupling of $\lambda$ with itself. At a paired location $(x,y)$,
$$
 |u(x)-v(x)|\le |u(x)-v(y)|+L_u|x-y|
 \le L_j\rho_\sfE(j_u(x),j_v(y)).
$$
Integration and minimization give $\int_\sfX |u-v|\,\dd\lambda\le L_j\delta$. Put $M_{uv}=\|u-v\|_\infty$ and $B_*=1+2L_u+\operatorname{diam}\sfV$. At a maximizer, $|u-v|\ge M_{uv}/2$ within distance $M_{uv}/(2B_*)\le1/2$. The intersection of that ball with the cube has volume bounded below by a dimensional constant times its radius to the power $d$. Thus
\begin{equation}\label{eq:s4-graph-lift}
 \|u-v\|_\infty\le C_G\delta^{1/(d+1)},
 \qquad C_G=C_G(d,L_u,\operatorname{diam}\sfV).
\end{equation}
For fixed $(\alpha,c)$, apply \eqref{eq:s4-cost-response} to the two functions $y\mapsto c(\gamma_u,z,j_u(y))$ and $y\mapsto c(\gamma_v,z',j_v(y))$. Couple their minimizing spatial measures, then change the graph lift at the same position. Equations~\eqref{eq:s4-costclass} and \eqref{eq:s4-graph-lift} yield
$$
\begin{aligned}
 &W_1(K_{\alpha,c}[\gamma_u,z],K_{\alpha,c}[\gamma_v,z'])\\
 &\quad\le L_j(L_c/m_{d,\tau})^{1/3}
       \bigl(W_1(\gamma_u,\gamma_v)+\rho_\sfE(z,z')\bigr)^{\beta/3}
       +C_G\delta^{1/(d+1)}.
\end{aligned}
$$
All input distances are bounded, so this is at most a constant times $(W_1(\gamma_u,\gamma_v)+\rho_\sfE(z,z'))^\eta$. At fixed $(\gamma_v,z')$, combine \eqref{eq:s4-cost-response}, the coefficient estimate \eqref{eq:s4-coefficient}, and the Lipschitz pushforward bound to control the change from $(\alpha,c)$ to $(\widetilde\alpha,\widetilde c)$. The triangle inequality with intermediate row $K_{\alpha,c}[\gamma_v,z']$ proves \eqref{eq:s4-graph-stability}.

The class $\mathcal U_{L_u}$ is compact in the uniform norm by Arzel\`a--Ascoli; $u\mapsto\gamma_u$ is continuous and injective, since $\lambda$ has full support and $u$ is continuous. Hence the set of admissible triples $(\gamma_u,z,j_u(y))$ is compact. The bounds in \eqref{eq:s4-costclass} make $\mathfrak C$ convex, closed, bounded, and equicontinuous there, so it is compact. Existence and uniqueness were proved in \eqref{eq:s4-uniform-growth}; pushing the spatial optimizer through $j_u$ gives the unique token row. Continuity implies measurability. Compactness is asserted for the cost--coefficient coordinates, not for uniform values of extended-valued regularizers at $\alpha=0$.\hfill$\square$

\subsection{Architecture examples}\label{app:s4-endpoints}

For graph contexts, write
$$
    F_\mu(\kappa):=\tfrac12W_2^2\bigl((\pi_\sfX)_\#\kappa,(\pi_\sfX)_\#\mu\bigr),
    \qquad \mathcal R_\alpha(\kappa,\mu):=(1-\alpha)F_\mu(\kappa)
                           +\alpha\tau\operatorname{KL}(\kappa\Vert\mu).
$$
Table~\ref{tab:variational-architectures} records the three architecture reads in this family and the separate Pearson construction from the representation theorem. The latter is proved in Section~\ref{app:variational-representation}; its decoder costs need not belong to the joint cost class $\mathfrak C$.

\begin{table}[t]
    \centering
    \small
    \begin{tabular}{@{}p{.19\linewidth}p{.32\linewidth}p{.41\linewidth}@{}}
        \toprule
        Read / architecture & Regularizer & Cost \\
        \midrule
        SelfWarp & $F_\mu(\kappa)$ & $B|\pi_\sfX(w)-a(\mu,z)|$, $B>\sqrt d$ \\[1ex]
        Attention & $\tau\operatorname{KL}(\kappa\Vert\mu)$ & $-\langle Qz,K_{\mathrm{key}}w\rangle$ \\[1ex]
        FNO uniform read & $\mathcal R_\alpha(\kappa,\mu)$ & $0$ \\[1ex]
        \midrule
        Representation theorem & $\tfrac12\chi^2(\kappa\Vert\mu)$ & $c_{jk}(\mu,z,w)=-\theta_j(\mu,z)\psi_k(w)$ \\
        \bottomrule
    \end{tabular}
    \caption{Variational read constructions. The first three rows belong to the Wasserstein--KL family on the stated graph class, with an admissible H\"older address $a(\mu,z)=\pi_\sfX(z)+g(\mu,z)\in\sfX$ for SelfWarp. The last row is the separate Pearson realization; its calibration head has zero cost.}
    \label{tab:variational-architectures}
    \label{tab:s4-architectures}
\end{table}

At $\alpha=1$, put $Z=\int_\sfX e^{-c/\tau}\,\dd\lambda$. Direct expansion gives
$$
 \int_\sfX  c\,\dd\nu+\tau H(\nu)
 =\tau\operatorname{KL}\!\left(\nu\,\middle\Vert\,Z^{-1}e^{-c/\tau}\lambda\right)-\tau\log Z.
$$
This proves the normalized attention row in Table~\ref{tab:s4-architectures}, including extended values outside the entropy domain. For the SelfWarp cost in the table, \eqref{eq:s4-F-Lip} gives
$$
\begin{aligned}
 &\int_\sfX  B|y-\pi_\sfX z-g(\gamma_u,z)|\,\nu(\dd y)+F(\nu)
      -F\!\left(\delta_{\pi_\sfX z+g(\gamma_u,z)}\right)\\
 &\hspace{2cm}\ge(B-D)W_1\!\left(\nu,\delta_{\pi_\sfX z+g(\gamma_u,z)}\right).
\end{aligned}
$$
Therefore $B>D$ suffices for the unique spatial optimizer $\delta_{\pi_\sfX z+g(\gamma_u,z)}$. It need not be necessary for a particular address. For zero cost, both $F$ and $H$ are nonnegative and vanish only at $\lambda$, so every $\alpha$ gives $K[\gamma_u,z]=\gamma_u$. At $\alpha=0$, the spatial optimizer also has the transport-map description \eqref{eq:s4-free-map}. The point-read identification agrees with Proposition~\ref{prop:wasserstein-selfwarp}.

A bounded Lipschitz attention score in $(z,w)$ is spatially Lipschitz after substitution $w=j_u(y)$. Equation~\eqref{eq:s4-graph-lift} controls its dependence on $u$, giving \eqref{eq:s4-costclass}, for example with $\beta=1/(d+1)$. For the warp costs,
$$
\begin{aligned}
 &\big|B|y-\pi_\sfX z-g(\gamma_u,z)|-B|y'-\pi_\sfX z'-g(\gamma_v,z')|\big|\\
 &\qquad\le B\bigl(|y-y'|+\rho_\sfE(z,z')+|g(\gamma_u,z)-g(\gamma_v,z')|\bigr).
\end{aligned}
$$
A H\"older displacement map therefore gives an admissible cost. Choose a common exponent no larger than $1/(d+1)$ and the displacement exponent, and common bounds covering both families. Then the endpoints and their straight interpolation belong to $\mathfrak C\times[0,1]$.

\subsection{Grid consistency and the obstruction from changing supports}\label{app:s4-discrete}
The graph theorem does not by itself certify optimization on empirical contexts. The following approximation statement supplies a separate numerical guarantee. Partition $\sfX$ into cubes $Q_i$ of side $h=1/q$, with centers $x_i$ and weights $w_i=q^{-d}$. Let $\lambda_h=\sum_iw_i\delta_{x_i}$. For a probability vector $p=(p_i)$, set $\nu_h(p)=\sum_ip_i\delta_{x_i}$ and
\begin{equation}\label{eq:s4-discrete-problem}
 J^h_{\alpha,\widehat c}(p)
 =\sum_i\widehat c_i p_i+\frac{1-\alpha}{2}W_2^2(\lambda_h,\nu_h(p))
       +\alpha\tau\sum_i p_i\log(p_i/w_i).
\end{equation}
This is a convex problem. Its entropy acts on the destination row, not on the transport coupling.

\paragraph{Uniform grid consistency.}
Suppose $\operatorname{Lip}(c)\le L_y$, $\max_i|\widehat c_i-c(x_i)|\le e_c$, and $p_h$ has objective gap at most $e_{\mathrm{solve}}$ in \eqref{eq:s4-discrete-problem}. Then, for all $\alpha\in[0,1]$,
\begin{equation}\label{eq:s4-grid-bound}
 W_1(\nu_h(p_h),\nu_{\alpha,c})
 \le Dh+\left[\frac{(3DL_y+5D^2)h+2e_c+e_{\mathrm{solve}}}{m_{d,\tau}}\right]^{1/3}.
\end{equation}
For exact samples of $u\in\mathcal U_{L_u}$, pushing the row to $j_u(x_i)$ multiplies this bound by at most $\sqrt{1+L_u^2}$.
\begin{proof}
Lift $p$ to the histogram $L_hp=\sum_i(p_i/w_i)\mathbf1_{Q_i}\lambda$. Its entropy equals the discrete entropy, and $W_1(L_hp,\nu_h(p))\le Dh$. Coupling source and destination points to their cell centers shows that the transport terms differ by at most $2D^2h$; the cost terms differ by at most $DL_yh+e_c$. Thus the continuum and discrete objectives differ by at most $C_1h+e_c$, where $C_1=DL_y+2D^2$.

Take as a discrete competitor the cell masses of $\nu_{\alpha,c}$. Its histogram is $S_q\nu_{\alpha,c}$, whose continuum objective is at most the minimum plus $C_0h$, with $C_0=DL_y+D^2$, by \eqref{eq:s4-histogram} and \eqref{eq:s4-F-Lip}. When $\alpha=0$ no entropy comparison is needed. Comparing the discrete optimizer to this competitor gives a continuum objective gap for $L_hp_h$ of at most
$$
 (C_0+2C_1)h+2e_c+e_{\mathrm{solve}}.
$$
Apply \eqref{eq:s4-uniform-growth} and add the lifting distance $Dh$. This proves \eqref{eq:s4-grid-bound}, for every approximate minimizer, including nonunique discrete minimizers at $\alpha=0$.
\end{proof}

\paragraph{Why non-collapse is insufficient.}
Take the ambient interval $\sfE=\sfX=[0,2]$, $\pi_\sfX=\mathrm{id}$, the fixed cost $c(w)=-3w$, and
$$
 \mu_\varepsilon=\frac{1-\varepsilon}{2}(\delta_0+\delta_1)+\varepsilon\delta_2,
 \qquad 0\le\varepsilon\le\tfrac14.
$$
These measures remain a positive distance from every Dirac measure and $W_1(\mu_\varepsilon,\mu_0)=3\varepsilon/2$. At $\alpha=0$, the source--destination objective is $-3y+|x-y|^2/2$, strictly decreasing in the available destination $y\in[0,2]$. Its unique minimizing row is therefore $\delta_2$ for $\varepsilon>0$, but $\delta_1$ for $\varepsilon=0$. Thus the input distance tends to zero while the distance between the minimizing reads remains one. The theorem therefore cannot extend to arbitrary non-collapsing contexts. The grid bound concerns a specified approximation of the continuum problem.

\end{document}

%% file: figures/operator-learning-input.tex
\begingroup
\definecolor{olIdeal}{HTML}{236B91}
\definecolor{olSample}{HTML}{B95720}
\definecolor{olAxis}{HTML}{8A9098}
\begin{tikzpicture}[
    x=1.15cm, y=1.25cm,
    font=\small,
    axis/.style={draw=olAxis, line width=.5pt, -{Stealth[length=3.5pt]}},
    curve/.style={line width=1.05pt, line cap=round, line join=round},
    every node/.style={inner sep=2pt},
    declare function={
        olinput(\x)=.50+.72*(1-cos(180*\x/4.3));
    }
]
    \path[use as bounding box] (0,0) rectangle (6.0,3.40);
    \begin{scope}[shift={(.28,.28)}]
        \draw[axis] (-.12,0) -- (4.85,0) node[right, text=black] {$\sfX$};
        \draw[axis] (0,-.10) -- (0,2.72)
            node[left, text=black] {$\sfY$};
        \draw[curve, olIdeal, domain=0:4.3, samples=65]
            plot (\x,{olinput(\x)});
        \node[olIdeal, anchor=west] at (4.35,1.96) {$f$};
        \foreach \x/\noise in {
            .22/.18, .72/-.16, 1.16/.24, 1.66/-.19,
            2.12/.16, 2.67/-.22, 3.17/.23, 3.72/-.15, 4.12/.10,
            .72/.12, 1.66/.09, 1.66/.37, 2.67/.08,
            3.17/-.05, 3.17/-.33, 3.72/.15
        } {
            \fill[olSample] (\x,{olinput(\x)+\noise}) circle[radius=1.8pt];
        }
        \node[olSample] at (4.30,1.60)
            {$\operatorname{supp}\mu$};
    \end{scope}
\end{tikzpicture}\endgroup

%% file: figures/operator-learning-output.tex
\begingroup
\definecolor{olIdeal}{HTML}{BFC3C9}
\definecolor{olSample}{HTML}{B95720}
\definecolor{olNetwork}{HTML}{236B91}
\definecolor{olAxis}{HTML}{8A9098}
\begin{tikzpicture}[
    x=1.15cm, y=1.25cm,
    font=\small,
    axis/.style={draw=olAxis, line width=.5pt, -{Stealth[length=3.5pt]}},
    curve/.style={line width=1.05pt, line cap=round, line join=round},
    every node/.style={inner sep=2pt},
    declare function={
        olbase(\x)=1.43+.45*cos(180*\x/4.3);
        olend(\x)=max(0,1-\x/.86)^2+max(0,1-(4.3-\x)/.86)^2;
        oltarget(\x)=olbase(\x)-.20*olend(\x)-.065*sin(180*\x/4.3)^2;
        olideal(\x)=olbase(\x)+.16*olend(\x);
        olsample(\x)=olideal(\x)+.18*cos(360*\x/4.3);
    }
]
    \path[use as bounding box] (0,0) rectangle (6.0,3.40);
    \begin{scope}[shift={(.28,.28)}]
        \draw[axis] (-.12,0) -- (4.85,0) node[right, text=black] {$\sfX$};
        \draw[axis] (0,-.10) -- (0,2.72)
            node[left, text=black] {$\sfY$};
        \draw[curve, olIdeal, domain=0:4.3, samples=101]
            plot (\x,{oltarget(\x)});
        \draw[curve, olNetwork, domain=0:4.3, samples=101]
            plot (\x,{olideal(\x)});
        \draw[curve, olSample, domain=0:4.3, samples=151]
            plot (\x,{olsample(\x)});
        \node[olSample, anchor=west, inner sep=1pt] at (4.30,1.50)
            {$\calN(\mu,\cdot)$};
        \node[olNetwork, anchor=west, inner sep=1pt] at (4.30,1.14)
            {$\calN(\gamma_f,\cdot)$};
        \node[olIdeal, anchor=west, inner sep=1pt] at (4.30,.78) {$A[f]$};
    \end{scope}
\end{tikzpicture}\endgroup

%% file: figures/non-admissible-measure.tex
\begingroup
\definecolor{naFunction}{HTML}{236B91}
\definecolor{naSample}{HTML}{B95720}
\definecolor{naAxis}{HTML}{8A9098}
\begin{tikzpicture}[
    x=1cm, y=1cm,
    font=\footnotesize,
    axis/.style={draw=naAxis, line width=.45pt, -{Stealth[length=3pt]}},
    every node/.style={inner sep=1.5pt},
    declare function={nafunction(\x)=.35+.50*(1-cos(180*\x/2.8));}
]
    \path[use as bounding box] (0,0) rectangle (4.05,2.40);
    \begin{scope}[shift={(.25,.32)},yscale=.92]
        \draw[axis] (-.10,0) -- (3.25,0) node[right, text=black] {$\sfX$};
        \draw[axis] (0,-.07) -- (0,1.90)
            node[left, text=black] {$\sfY$};
        \draw[naAxis!65, densely dotted, line width=.45pt]
            (1.20,0) -- (1.20,1.76);
        \draw[naAxis, line width=.45pt] (1.20,.035) -- (1.20,-.035)
            node[below, text=black] {$a$};
        \draw[naFunction, line width=.9pt, line cap=round,
            domain=0:2.8, samples=65] plot (\x,{nafunction(\x)});
        \node[naFunction, anchor=west] at (2.84,1.36) {$f$};
        \foreach \y in {.18,.38,.50,.66,.78,.92,1.12,1.40,1.70} {
            \fill[naSample] (1.20,\y) circle[radius=1.5pt];
        }
        \node[naSample, anchor=west] at (1.27,1.83) {$\operatorname{supp}\mu$};
    \end{scope}
\end{tikzpicture}\endgroup

%% file: references.bib
@misc{toscano2026athenaagenticteamhierarchical,
      title={ATHENA: Agentic Team for Hierarchical Evolutionary Numerical Algorithms}, 
      author={Juan Diego Toscano and Daniel T. Chen and George Em Karniadakis},
      year={2026},
      eprint={2512.03476},
      archivePrefix={arXiv},
      primaryClass={cs.LG},
      url={https://arxiv.org/abs/2512.03476}, 
}

@inproceedings{muser2026flowers,
  title = {Flowers: A Warp Drive for Neural {PDE} Solvers},
  author = {Muser, Till and Spitzer, Alexandra and Lassas, Matti and {de Hoop}, Maarten V. and Dokmani{\'c}, Ivan},
  year = {2026},
  eprint = {2603.04430},
  archivePrefix = {arXiv},
  primaryClass = {cs.LG},
  url = {https://arxiv.org/abs/2603.04430v2},
  note = {Version 2, 1 June 2026},
  booktitle = {Forty-Third International Conference on Machine Learning}
}

@inproceedings{furuya2024transformersInContext,
  title = {Transformers Are Universal In-context Learners},
  author = {Furuya, Takashi and {de Hoop}, Maarten V. and Peyr{\'e}, Gabriel},
  booktitle = {The Thirteenth International Conference on Learning Representations},
  year = {2025},
  note = {arXiv:2408.01367},
  eprint = {2408.01367},
  archivePrefix = {arXiv},
  primaryClass = {cs.CL},
  url = {https://arxiv.org/abs/2408.01367v2}
}

@inproceedings{vaswani2017attention,
  title = {Attention Is All You Need},
  author = {Vaswani, Ashish and Shazeer, Noam and Parmar, Niki and Uszkoreit, Jakob and Jones, Llion and Gomez, Aidan N. and Kaiser, {\L}ukasz and Polosukhin, Illia},
  booktitle = {Advances in Neural Information Processing Systems},
  pages = {5998--6008},
  year = {2017},
  volume = {30},
  publisher = {Curran Associates, Inc.},
  eprint = {1706.03762},
  archivePrefix = {arXiv},
  primaryClass = {cs.CL},
  url = {https://arxiv.org/abs/1706.03762v7}
}

@misc{furuya2026functiongraph,
  title = {Function Graph Transformers Universally Approximate Operators Between Function Spaces},
  author = {Furuya, Takashi and Mis, David and Dokmani{\'c}, Ivan and {de Hoop}, Maarten V. and Lassas, Matti},
  year = {2026},
  eprint = {2605.17968},
  archivePrefix = {arXiv},
  primaryClass = {cs.LG},
  url = {https://arxiv.org/abs/2605.17968v1}
}

@article{chen1995operators,
  title = {Universal Approximation to Nonlinear Operators by Neural Networks with Arbitrary Activation Functions and Its Application to Dynamical Systems},
  author = {Chen, Tianping and Chen, Hong},
  journal = {IEEE Transactions on Neural Networks},
  volume = {6},
  number = {4},
  pages = {911--917},
  year = {1995},
  doi = {10.1109/72.392253}
}

@book{villani2009optimal,
  title = {Optimal Transport: Old and New},
  author = {Villani, C{\'e}dric},
  series = {Grundlehren der mathematischen Wissenschaften},
  volume = {338},
  publisher = {Springer},
  address = {Berlin},
  year = {2009},
  doi = {10.1007/978-3-540-71050-9}
}

@misc{furuya2025support,
  title = {Transformers through the Lens of Support-Preserving Maps between Measures},
  author = {Furuya, Takashi and {de Hoop}, Maarten V. and Lassas, Matti},
  year = {2025},
  eprint = {2509.25611},
  archivePrefix = {arXiv},
  primaryClass = {cs.CL},
  doi = {10.48550/arXiv.2509.25611},
  url = {https://arxiv.org/abs/2509.25611v1},
  note = {Version 1, 30 September 2025}
}

@inproceedings{castin2024smooth,
  title = {{How Smooth Is Attention?}},
  author = {Castin, Val{\'e}rie and Ablin, Pierre and Peyr{\'e}, Gabriel},
  year = {2024},
  booktitle = {Proceedings of the 41st International Conference on Machine Learning},
  eprint = {2312.14820},
  archivePrefix = {arXiv},
  primaryClass = {cs.LG},
  url = {https://arxiv.org/abs/2312.14820v2}
}

@inproceedings{li2021fourier,
  title = {{Fourier Neural Operator for Parametric Partial Differential Equations}},
  author = {Li, Zongyi and Kovachki, Nikola and Azizzadenesheli, Kamyar and Liu, Burigede and Bhattacharya, Kaushik and Stuart, Andrew and Anandkumar, Anima},
  year = {2021},
  booktitle = {International Conference on Learning Representations},
  eprint = {2010.08895},
  archivePrefix = {arXiv},
  primaryClass = {cs.LG},
  url = {https://arxiv.org/abs/2010.08895v3}
}

@article{kovachki2023neuraloperator,
  title = {{Neural Operator: Learning Maps Between Function Spaces}},
  author = {Kovachki, Nikola and Li, Zongyi and Liu, Burigede and Azizzadenesheli, Kamyar and Bhattacharya, Kaushik and Stuart, Andrew and Anandkumar, Anima},
  year = {2023},
  journal = {Journal of Machine Learning Research},
  volume = {24},
  number = {89},
  pages = {1--97},
  eprint = {2108.08481},
  archivePrefix = {arXiv},
  primaryClass = {cs.LG},
  url = {https://arxiv.org/abs/2108.08481v6}
}

@misc{lu2021deeponet,
  title = {{DeepONet: Learning nonlinear operators for identifying differential equations based on the universal approximation theorem of operators}},
  author = {Lu, Lu and Jin, Pengzhan and Karniadakis, George Em},
  year = {2019},
  eprint = {1910.03193},
  archivePrefix = {arXiv},
  primaryClass = {cs.LG},
  url = {https://arxiv.org/abs/1910.03193v3},
  note = {Preprint}
}

@misc{balaji2026hybrid,
  title = {{Hybrid operator learning of wave scattering maps in high-contrast media}},
  author = {Balaji, Advait and Teolis, Trevor and Mis, S. David and Lara Benitez, Jose Antonio and Wang, Chao and {de Hoop}, Maarten V.},
  year = {2026},
  eprint = {2602.11197},
  archivePrefix = {arXiv},
  primaryClass = {eess.SP},
  url = {https://arxiv.org/abs/2602.11197v1},
  note = {Preprint}
}

@inproceedings{flowers,
  title = {Flowers: A Warp Drive for Neural {PDE} Solvers},
  author = {Muser, Till and Spitzer, Alexandra and Lassas, Matti and {de Hoop}, Maarten V. and Dokmani{\'c}, Ivan},
  year = {2026},
  eprint = {2603.04430},
  archivePrefix = {arXiv},
  primaryClass = {cs.LG},
  url = {https://arxiv.org/abs/2603.04430v2},
  booktitle = {Forty-Third International Conference on Machine Learning},
  note = {arXiv:2603.04430}
}

@inproceedings{bartolucci2023representation,
  author = {Bartolucci, Francesca and de B\'ezenac, Emmanuel and Raoni\'c, Bogdan and Molinaro, Roberto and Mishra, Siddhartha and Alaifari, Rima},
  title = {Representation Equivalent Neural Operators: a Framework for Alias-free Operator Learning},
  booktitle = {Advances in Neural Information Processing Systems},
  volume = {36}, year = {2023},
  eprint = {2305.19913}, archivePrefix = {arXiv},
  url = {https://arxiv.org/abs/2305.19913v2}
}

@misc{bendahi2026discretization,
  author = {Bendahi, Abderrahim and Fradin, Adrien and Peralez, Johan and Digne, Julie and Nadri, Madiha},
  title = {Stability and Discretization Error of State Space Model Neural Operators},
  year = {2026}, eprint = {2605.18905}, archivePrefix = {arXiv},
  url = {https://arxiv.org/abs/2605.18905v1}
}

@article{unser2019representer,
  author = {Unser, Michael},
  title = {A Representer Theorem for Deep Neural Networks},
  journal = {Journal of Machine Learning Research},
  year = {2019},
  volume = {20},
  number = {110},
  pages = {1--30},
  url = {https://www.jmlr.org/papers/v20/18-418.html}
}

@article{decarvalho2013euclidean,
  author = {de Carvalho, Miguel and Oumow, Boris and Segers, Johan and Warcho{\l}, Micha{\l}},
  title = {A {Euclidean} Likelihood Estimator for Bivariate Tail Dependence},
  journal = {Communications in Statistics---Theory and Methods},
  year = {2013},
  volume = {42},
  number = {7},
  pages = {1176--1192},
  doi = {10.1080/03610926.2012.709905},
  eprint = {1204.3524},
  archivePrefix = {arXiv},
  url = {https://arxiv.org/abs/1204.3524}
}

@inproceedings{niculae2017regularized,
  author = {Niculae, Vlad and Blondel, Mathieu},
  title = {A Regularized Framework for Sparse and Structured Neural Attention},
  booktitle = {Advances in Neural Information Processing Systems},
  year = {2017},
  volume = {30},
  pages = {3338--3348},
  publisher = {Curran Associates, Inc.},
  url = {https://proceedings.neurips.cc/paper/2017/hash/2d1b2a5ff364606ff041650887723470-Abstract.html}
}

@article{martins2022sparsecontinuous,
  author = {Martins, Andr{\'e} F. T. and Treviso, Marcos and Farinhas, Ant{\'o}nio and Aguiar, Pedro M. Q. and Figueiredo, M{\'a}rio A. T. and Blondel, Mathieu and Niculae, Vlad},
  title = {Sparse Continuous Distributions and {Fenchel--Young} Losses},
  journal = {Journal of Machine Learning Research},
  year = {2022},
  volume = {23},
  number = {257},
  pages = {1--74},
  url = {https://jmlr.org/papers/v23/21-0879.html}
}

@article{thorpe2017transportation,
  title = {A Transportation {$L^p$} Distance for Signal Analysis},
  author = {Thorpe, Matthew and Park, Serim and Kolouri, Soheil and Rohde, Gustavo K. and Slep{\v c}ev, Dejan},
  journal = {Journal of Mathematical Imaging and Vision},
  volume = {59},
  number = {2},
  pages = {187--210},
  year = {2017},
  doi = {10.1007/s10851-017-0726-4},
  url = {https://pmc.ncbi.nlm.nih.gov/articles/PMC6141213/},
  eprint = {1609.08669},
  archivePrefix = {arXiv},
  primaryClass = {cs.CV}
}

@article{garciatrillos2016continuum,
  author = {Garc{\'i}a Trillos, Nicol{\'a}s and Slep{\v c}ev, Dejan},
  title = {Continuum limit of total variation on point clouds},
  journal = {Archive for Rational Mechanics and Analysis},
  volume = {220},
  pages = {193--241},
  year = {2016},
  doi = {10.1007/s00205-015-0929-z},
  eprint = {1403.6355},
  archivePrefix = {arXiv},
  primaryClass = {math.ST},
  url = {https://arxiv.org/abs/1403.6355v3}
}

@inproceedings{cortes2017adanet,
  title={{AdaNet}: Adaptive Structural Learning of Artificial Neural Networks},
  author={Cortes, Corinna and Gonzalvo, Xavier and Kuznetsov, Vitaly and Mohri, Mehryar and Yang, Scott},
  booktitle={Proceedings of the 34th International Conference on Machine Learning},
  series={Proceedings of Machine Learning Research}, volume={70}, pages={874--883}, year={2017},
  url={https://proceedings.mlr.press/v70/cortes17a.html}
}

@inproceedings{martins2016sparsemax,
  author = {Martins, Andr{\'e} F. T. and Astudillo, Ram{\'o}n F.},
  title = {From Softmax to Sparsemax: A Sparse Model of Attention and Multi-Label Classification},
  booktitle = {Proceedings of the 33rd International Conference on Machine Learning},
  editor = {Balcan, Maria Florina and Weinberger, Kilian Q.},
  series = {Proceedings of Machine Learning Research},
  volume = {48},
  pages = {1614--1623},
  year = {2016},
  publisher = {PMLR},
  url = {https://proceedings.mlr.press/v48/martins16.html}
}

@inproceedings{terjek2021moreau,
  title = {Moreau-Yosida $f$-divergences},
  author = {Terj{\'e}k, D{\'a}vid},
  booktitle = {Proceedings of the 38th International Conference on Machine Learning},
  pages = {10214--10224},
  year = {2021},
  editor = {Meila, Marina and Zhang, Tong},
  volume = {139},
  series = {Proceedings of Machine Learning Research},
  publisher = {PMLR},
  url = {https://proceedings.mlr.press/v139/terjek21a.html}
}

@inproceedings{debie2019stochastic,
  title = {Stochastic Deep Networks},
  author = {de Bie, Gwendoline and Peyr{\'e}, Gabriel and Cuturi, Marco},
  booktitle = {Proceedings of the 36th International Conference on Machine Learning},
  series = {Proceedings of Machine Learning Research},
  volume = {97},
  pages = {1556--1565},
  year = {2019},
  url = {https://proceedings.mlr.press/v97/de-bie19a.html}
}

@article{himmelberg1969selection,
  author = {Himmelberg, C. J. and Van Vleck, F. S.},
  title = {Some Selection Theorems for Measurable Functions},
  journal = {Canadian Journal of Mathematics},
  volume = {21},
  pages = {394--399},
  year = {1969},
  doi = {10.4153/CJM-1969-041-7},
  url = {https://www.cambridge.org/core/journals/canadian-journal-of-mathematics/article/some-selection-theorems-for-measurable-functions/30345D24D3DFFAAF9CE805C30EE959C6}
}

@article{dambrogi1999discretization,
  author = {D'Ambrogi, Barbara and M{\"a}enp{\"a}{\"a}, Sari and Markkanen, Markku},
  title = {Discretization Independent Retrieval of Atmospheric Ozone Profile},
  journal = {Geophysica},
  year = {1999},
  volume = {35},
  number = {1-2},
  pages = {87--99},
  url = {https://archive.geophysica.fi/pdf/geophysica_1999_35_1-2_087_dambrogi.pdf}
}

@article{lassas2004totalvariation,
  author = {Lassas, Matti and Siltanen, Samuli},
  title = {Can one use total variation prior for edge-preserving {Bayesian} inversion?},
  journal = {Inverse Problems},
  year = {2004},
  volume = {20},
  number = {5},
  pages = {1537--1563},
  doi = {10.1088/0266-5611/20/5/013},
  url = {https://siltanen-research.net/wp-content/uploads/publications/LassasSiltanen2004.pdf}
}

@article{lassas2009discretization,
  author = {Lassas, Matti and Saksman, Eero and Siltanen, Samuli},
  title = {Discretization-invariant {Bayesian} inversion and {Besov} space priors},
  journal = {Inverse Problems and Imaging},
  year = {2009},
  volume = {3},
  number = {1},
  pages = {87--122},
  doi = {10.3934/ipi.2009.3.87},
  url = {https://www.aimsciences.org/article/doi/10.3934/ipi.2009.3.87},
  eprint = {0901.4220},
  archivePrefix = {arXiv},
  primaryClass = {math.PR}
}

@article{stuart2010inverse,
  author = {Stuart, Andrew M.},
  title = {Inverse problems: A {Bayesian} perspective},
  journal = {Acta Numerica},
  year = {2010},
  volume = {19},
  pages = {451--559},
  doi = {10.1017/S0962492910000061},
  url = {https://doi.org/10.1017/S0962492910000061}
}

@phdthesis{lasanen2002discretizations,
  author = {Lasanen, Sari},
  title = {Discretizations of Generalized Random Variables with Applications to Inverse Problems},
  school = {University of Oulu},
  year = {2002},
  series = {Annales Academiae Scientiarum Fennicae, Mathematica Dissertationes},
  number = {130},
  publisher = {Suomalainen Tiedeakatemia},
  address = {Helsinki},
  isbn = {951-41-0919-8},
  note = {64 pages},
  url = {https://www.finna.fi/Record/utu.999531305405971}
}

@article{lehtinen1989linear,
  author = {Lehtinen, M. S. and P{\"a}iv{\"a}rinta, L. and Somersalo, E.},
  title = {Linear inverse problems for generalised random variables},
  journal = {Inverse Problems},
  year = {1989},
  volume = {5},
  number = {4},
  pages = {599--612},
  doi = {10.1088/0266-5611/5/4/011},
  url = {https://doi.org/10.1088/0266-5611/5/4/011}
}

@article{wang2023discretization,
  author = {Wang, Clinton J. and Golland, Polina},
  title = {Discretization Invariant Networks for Learning Maps between Neural Fields},
  journal = {Transactions on Machine Learning Research},
  year = {2023},
  url = {https://openreview.net/forum?id=CpYBAqDgmz},
  eprint = {2206.01178},
  archivePrefix = {arXiv},
  primaryClass = {cs.LG}
}

@inproceedings{ruiz2020graphon,
  author = {Ruiz, Luana and Chamon, Luiz F. O. and Ribeiro, Alejandro},
  title = {Graphon Neural Networks and the Transferability of Graph Neural Networks},
  booktitle = {Advances in Neural Information Processing Systems},
  volume = {33},
  year = {2020},
  url = {https://papers.nips.cc/paper/2020/hash/12bcd658ef0a540cabc36cdf2b1046fd-Abstract.html}
}

@misc{lanthaler2024discretization,
  title = {Discretization Error of Fourier Neural Operators},
  author = {Lanthaler, Samuel and Stuart, Andrew M. and Trautner, Margaret},
  year = {2024},
  eprint = {2405.02221},
  archivePrefix = {arXiv},
  primaryClass = {math.NA},
  doi = {10.48550/arXiv.2405.02221},
  url = {https://arxiv.org/abs/2405.02221},
  note = {Version 2, 25 September 2025}
}

@misc{subedi2025controlling,
  title = {Controlling Statistical, Discretization, and Truncation Errors in Learning Fourier Linear Operators},
  author = {Subedi, Unique and Tewari, Ambuj},
  year = {2025},
  eprint = {2408.09004},
  archivePrefix = {arXiv},
  primaryClass = {stat.ML},
  doi = {10.48550/arXiv.2408.09004},
  url = {https://arxiv.org/abs/2408.09004},
  note = {Version 2, 6 February 2025; first posted in 2024 as Error Bounds for Learning Fourier Linear Operators}
}

@inproceedings{gao2025discretization,
  author = {Gao, Wenhan and Xu, Ruichen and Deng, Yuefan and Liu, Yi},
  title = {Discretization-invariance? On the Discretization Mismatch Errors in Neural Operators},
  booktitle = {International Conference on Learning Representations},
  year = {2025},
  url = {https://proceedings.iclr.cc/paper_files/paper/2025/hash/313829757739365201b5adb3a1cbd9bd-Abstract-Conference.html}
}

@inproceedings{roulet2025fdivergences,
  title = {Loss Functions and Operators Generated by $f$-Divergences},
  author = {Roulet, Vincent and Liu, Tianlin and Vieillard, Nino and Sander, Micha{\"e}l E. and Blondel, Mathieu},
  booktitle = {Proceedings of the 42nd International Conference on Machine Learning},
  pages = {52110--52138},
  year = {2025},
  volume = {267},
  series = {Proceedings of Machine Learning Research},
  publisher = {PMLR},
  url = {https://proceedings.mlr.press/v267/roulet25a.html}
}
